\documentclass[10pt,twocolumn,letterpaper]{article}

\usepackage[section]{placeins}
\usepackage[pagenumbers]{cvpr} % To force page numbers, e.g. for an arXiv version

\usepackage{fancyhdr}
\usepackage{tabularray}
\usepackage{float}
\usepackage[normalem]{ulem}
\usepackage[dvipsnames]{xcolor}
\usepackage[accsupp]{axessibility} % Improves PDF readability for those with visual impairments.

\definecolor{cvprblue}{rgb}{0.21,0.49,0.74}
\usepackage[pagebackref,breaklinks,colorlinks,citecolor=cvprblue]{hyperref}

\def\paperID{8266} % *** Enter the Paper ID here
\def\confName{CVPR}
\def\confYear{2024}

\title{Content-Adaptive Non-Local Convolution for Remote Sensing Pansharpening}

\makeatletter
\def\@fnsymbol#1{\ensuremath{\ifcase#1\or \dagger\or *\or\ddagger\or
   \mathsection\or \mathparagraph\or \|\or **\or \dagger\dagger
   \or \ddagger\ddagger \else\@ctrerr\fi}}
\makeatother

\author{Yule Duan\thanks{Contributed Equally.}, Xiao Wu\footnotemark[1], Haoyu Deng, Liang-Jian Deng\thanks{Corresponding author.}\\
University of Electronic Science and Technology of China, China\\
{\tt\small yule.duan@outlook.com; wxwsx1997@gmail.com;} \\ {\tt\small haoyu\_deng@std.uestc.edu.cn; liangjian.deng@uestc.edu.cn}
}

\AddToShipoutPicture*{
    \color[rgb]{.5,.5,.5}
    \AtTextUpperLeft{%confidential
      \put(0,\LenToUnit{1cm}){\parbox{\textwidth}{\copyright 2024 IEEE.  Personal use of this material is permitted.  Permission from IEEE must be obtained for all other uses, in any current or future media, including reprinting/republishing this material for advertising or promotional purposes, creating new collective works, for resale or redistribution to servers or lists, or reuse of any copyrighted component of this work in other works.}}
    }
}

\begin{document}
\maketitle
\begin{abstract}
Currently, machine learning-based methods for remote sensing pansharpening have progressed rapidly. However, existing pansharpening methods often do not fully exploit differentiating regional information in non-local spaces, thereby limiting the effectiveness of the methods and resulting in redundant learning parameters. In this paper, we introduce a so-called content-adaptive non-local convolution (CANConv), a novel method tailored for remote sensing image pansharpening. Specifically, CANConv employs adaptive convolution, ensuring spatial adaptability, and incorporates non-local self-similarity through the similarity relationship partition (SRP) and the partition-wise adaptive convolution (PWAC) sub-modules. Furthermore, we also propose a corresponding network architecture, called CANNet, which mainly utilizes the multi-scale self-similarity. Extensive experiments demonstrate the superior performance of CANConv, compared with recent promising fusion methods. Besides, we substantiate the method's effectiveness through visualization, ablation experiments, and comparison with existing methods on multiple test sets. The source code is publicly available at \url{https://github.com/duanyll/CANConv}.
\end{abstract} 
\section{Introduction}

\begin{figure}[t]
   \centering
   \includegraphics[width=1.0\linewidth]{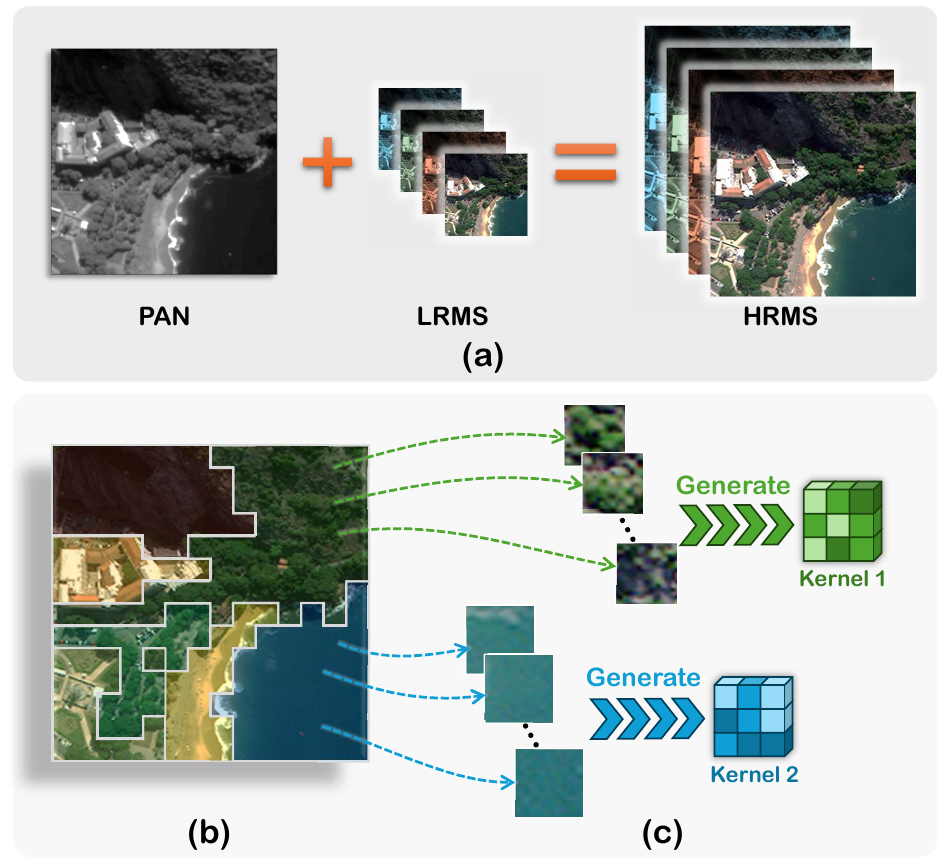}

   \caption{(a) Pansharpening involves fusing the PAN and LRMS images into an HRMS image. (b) A toy example of partitioned regions and (c) their corresponding content-adaptive convolution kernels, which is the motivation of this paper: 1) different content (regions) should be filtered by distinct kernels; 2) non-local information with the same content (regions) is extracted and represented only by the same convolution kernel. }
   \label{fig:Intro}
   \vspace{-0.5cm}
\end{figure}

Due to technological constraints, existing remote sensing satellites can only capture low-resolution multispectral images (LRMS) and high-resolution panchromatic images (PAN). Wherein, LRMS images consist of four, eight, or more channels in different bands, but have low spatial resolution. PAN images usually have 4$\times$ higher spatial resolution, but they are monochromatic and grayscale. Especially, 
pansharpening in this work involves merging an LRMS image and a PAN image to produce a high-resolution multispectral (HRMS) image, see \cref{fig:Intro}. Also shown in \cref{fig:Intro}, remote sensing images possess unique characteristics compared to common natural images. Just as facial images are composed of fixed parts such as eyes, nose, and mouth, remote sensing images also consist of relatively stable elements, such as oceans, forests, buildings and streets, \etc. These components exhibit distinct and easily distinguishable features in terms of color and texture.  Within regions with similar semantics, there are numerous repetitive tiled textures, and even in distant locations, similar textures can be found. Given these characteristics of remote sensing images, an ideal pansharpening method should be able to adapt to different regions with varying features and should leverage information from non-local similar regions.

To obtain HRMS images, various prior arts have been proposed for the task of pansharpening. These methods include traditional methods and modern deep learning-based methods. Specifically, traditional pansharpening methods can be categorized into three types~\cite{mengReviewPansharpeningMethods2019}: component substitution (CS) methods~\cite{choiNewAdaptiveComponentSubstitutionBased2011,vivoneRobustBandDependentSpatialDetail2019}, multi-resolution analysis (MRA) methods~\cite{vivoneContrastErrorBasedFusion2014,vivoneFullScaleRegressionBased2018}, and variational optimization (VO)-based methods~\cite{fuVariationalPanSharpeningLocal2019,tianVariationalPansharpeningExploiting2022}. In recent years, a plethora of methods based on convolutional neural networks (CNN) has also been employed for pansharpening, such as DiCNN~\cite{hePansharpeningDetailInjection2019}, PanNet~\cite{yangPanNetDeepNetwork2017}, and FusionNet~\cite{dengDetailInjectionBasedDeep2021}, \etc. Compared to traditional approaches, CNN-based methods have made significant progress in performance.

\begin{figure}
   \centering
   \includegraphics[width=1.0\linewidth]{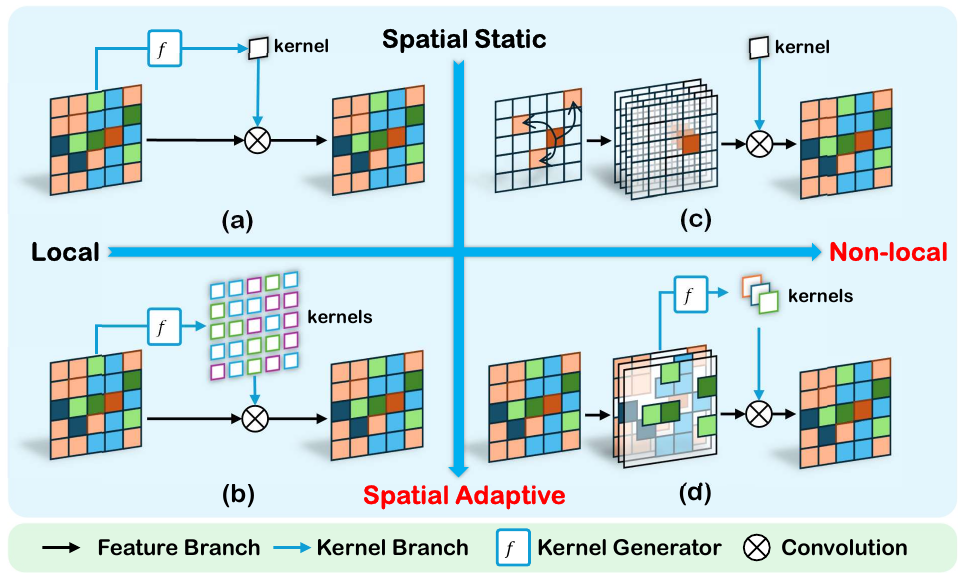}

   \caption{Overall workflow of four convolution methods related to adaptivity and non-locality. (a) Global adaptive/standard convolution~\cite{jiaDynamicFilterNetworks2016}. (b) Spatial adaptive convolution~\cite{suPixelAdaptiveConvolutionalNeural2019, zhouDecoupledDynamicFilter2021, jinLAGConvLocalContextAdaptive2022}. (c) Graph convolution~\cite{liCrossPatchGraphConvolutional2021, zhouCrossScaleInternalGraph2020}. (d) The proposed method.}
   \label{fig:FourMethods}
   \vspace{-0.5cm}
\end{figure}

Among CNN-based pansharpening methods, in particular, local content-adaptive convolution techniques deserve special attention. As depicted in \cref{fig:FourMethods}, an early adaptive convolution method, namely DFN~\cite{jiaDynamicFilterNetworks2016}, resembles standard convolution operators, employing spatially static convolution kernels that cannot adapt to differences in various parts of the space. However, DFN dynamically generates convolution kernels based on the input content through an extra network branch, enabling it to adapt to input variations and provide better feature extraction capabilities than standard convolution. Subsequently proposed spatial adaptive convolution methods, such as PAC~\cite{suPixelAdaptiveConvolutionalNeural2019}, DDF~\cite{zhouDecoupledDynamicFilter2021}, and LAGConv~\cite{jinLAGConvLocalContextAdaptive2022}, \etc, generate different convolution kernels for each pixel based on the input, allowing them to adapt to differences in different spatial regions. However, these approaches have an obvious drawback: generating convolution kernels for each pixel consumes a considerable amount of redundant computation power and memory. Also, they generate convolution kernels based on local information and lack the ability to utilize non-local information. Graph convolution methods, such as CPNet~\cite{liCrossPatchGraphConvolutional2021} and IGNN~\cite{zhouCrossScaleInternalGraph2020}, have the ability to leverage non-local information. IGNN models self-similarity relationships in the input using the k-NN algorithm, concatenating the most similar patches in the global scope after each patch in the channel dimension. It then employs standard convolution layers to process this additional similarity information. This approach can utilize limited information from non-local similar regions, but lacks spatial adaptability. Deformable Convolution~\cite{daiDeformableConvolutionalNetworks2017,zhuDeformableConvNetsV22018} can dynamically change the offsets of sampling points, but their convolution kernel weights cannot be changed dynamically, and it cannot purposefully obtain long-range texture information from dispersed sampling points.

To overcome the limitations of previous methods, we designed the Content-Adaptive Non-Local Convolution (CANConv) method tailored to the characteristics of remote sensing images in pansharpening tasks. CANConv achieves spatial adaptability through content-adaptive convolution and utilizes non-local self-similarity information, incorporating two sub-modules: Similarity Relationship Partition (SRP) and Partition-Wise Adaptive Convolution (PWAC). The main contributions of this paper are as follows:

\begin{enumerate}
    \item We propose the CANConv module, which utilizes adaptive convolution to simultaneously incorporate spatial adaptability and non-local self-similarity. Building upon the CANConv module, we design the CANNet network, capable of leveraging multi-scale self-similarity information.
    \item We analyze the prevalent non-local self-similarity relationships in remote sensing images. The theoretical effectiveness of the CANConv module is demonstrated through visualization and discussion experiments.
    \item We validate the CANConv method on multiple pansharpening datasets by comparing it with various pansharpening methods. The results indicate that CANConv achieves state-of-the-art performance.
\end{enumerate}

\section{Related Works}

\subsection{Content-Adaptive Convolution}

Compared to standard convolution operators that use global static convolution kernels, adaptive convolution operators employ different convolution kernels based on varying inputs, offering superior feature extraction capabilities and flexibility. Early adaptive convolution work, such as DFN~\cite{jiaDynamicFilterNetworks2016}, utilized a separate branch to generate convolution kernel filters. Subsequent efforts employed more intricate methods for convolution kernel generation. For instance, DYConv~\cite{chenDynamicConvolutionAttention2019} aggregated multiple convolution kernels using attention. These methods use the same kernels to filter different spatial regions, lacking the ability to adapt to distinct areas. DRConv~\cite{chenDynamicRegionAwareConvolution2021} considered spatial differences by selecting convolution kernels through independent convolution branches, yet it lacked adaptability. Spatial adaptive convolution methods apply a unique set of convolution kernels for each pixel, accommodating spatial differences in the input. PAC~\cite{suPixelAdaptiveConvolutionalNeural2019} fine-tuned convolution kernels using fixed Gaussian kernels for each pixel, but its flexibility was constrained. DDF~\cite{zhouDecoupledDynamicFilter2021} decoupled spatial adaptive convolution kernels in spatial and channel dimensions, somewhat reducing computational overhead but still faced redundancy due to generating a vast number of convolution kernels equal to the number of pixels. Involution~\cite{liInvolutionInvertingInherence2021} revealed the intrinsic connection between adaptive convolution and self-attention mechanisms.

Adaptive convolution methods designed specifically for remote sensing pansharpening tasks have already been developed. LAGConv~\cite{jinLAGConvLocalContextAdaptive2022} enhanced the ability of spatial adaptive convolution to gather information from local context, while ADKNet~\cite{pengSourceAdaptiveDiscriminativeKernels2022} tackled differences between PAN and LRMS images using distinct kernel generation branches. The success of these methods underscores the need to customize adaptive convolution techniques based on the characteristics of pansharpening tasks. Remote sensing images contain abundant non-local self-similarity information. However, in previous spatial adaptive convolution methods, the ability to utilize this information is absent.

\subsection{Non-Local Methods}

Many image restoration methods leverage information from non-local similar regions within images. Repeated patterns are prevalent in natural images, especially in remote sensing images. Traditional approaches like non-local means~\cite{buadesNonlocalAlgorithmImage2005} and BM3D~\cite{dabovImageDenoisingSparse2007} directly aggregate similar parts within images to denoise the image. Methods proposed by Mairal \etal~\cite{mairalNonlocalSparseModels2009} and Lecouat \etal~\cite{lecouatFullyTrainableInterpretable2020} exploit non-local self-similarity together with sparsity. The use of the k-nearest neighbors (kNN) algorithm is a vital technique for modeling non-local similarity relationships in images. Plötz and Roth~\cite{plotzNeuralNearestNeighbors2018} introduced a method to make the kNN algorithm differentiable in deep neural networks. Graph convolution networks based on the kNN algorithm, such as CPNet~\cite{liCrossPatchGraphConvolutional2021} and IGNN~\cite{zhouCrossScaleInternalGraph2020}, have shown significant effectiveness in image denoising and super-resolution tasks. In these methods, for each patch, $k$ most similar patches are identified to construct a similarity graph, and the found similar patches are concatenated with the original patch along the channel dimension. However, since increasing $k$ will bring significant growth in the number of parameters, the value of $k$ is usually small (less than 3), limiting the network's access to a very limited set of similar patches and preventing it from capturing the complete context of similarity. This limitation hampers the network to fully utilize non-local information.

\subsection{Motivation}

Remote sensing images contain distinct regions with different semantics, each characterized by unique and fixed features. Using the same convolution kernel to filter regions with different contents is not the most rational approach. Spatial adaptive convolution methods achieve adaptability to different regions by generating a large number of redundant convolution kernels. However, they cannot utilize information from similar patches that are spatially distant. To adapt to the distinct characteristics of different regions while comprehensively leveraging non-local self-similarity information in a global scope, we adopted a different approach from graph convolution. \textit{We modeled the self-similarity relationships in the image by clustering, rather than using nearest neighbors.} Our method first clusters pixels on the image based on the features of their neighboring regions. Therefore, we may assign each pixel to one cluster and an unlimited amount of similar pixels from anywhere in the image may form a single cluster. Subsequently, we aggregate all contents belonging to the same cluster and generate a set of convolution kernels adaptively for each cluster based on its contents. Finally, the same adaptive convolution kernel is applied to all pixels within each cluster, allowing information from all similar patches to be propagated through convolutional kernels. Since the number of clusters is much smaller than the number of pixels, \textit{our method significantly reduces the redundant computation required for generating convolution kernels} compared to common spatial adaptive convolution methods. Additionally, each cluster in our method contains a much larger number of pixels than the nearest neighbor patches aggregated by graph convolution methods, and the cluster number is not limited by the number of parameters, \textit{enabling our approach to acquire self-similar information more comprehensively}.

\section{Methods}

\begin{figure*}[t]
  \centering
  \includegraphics[width=1.0\linewidth]{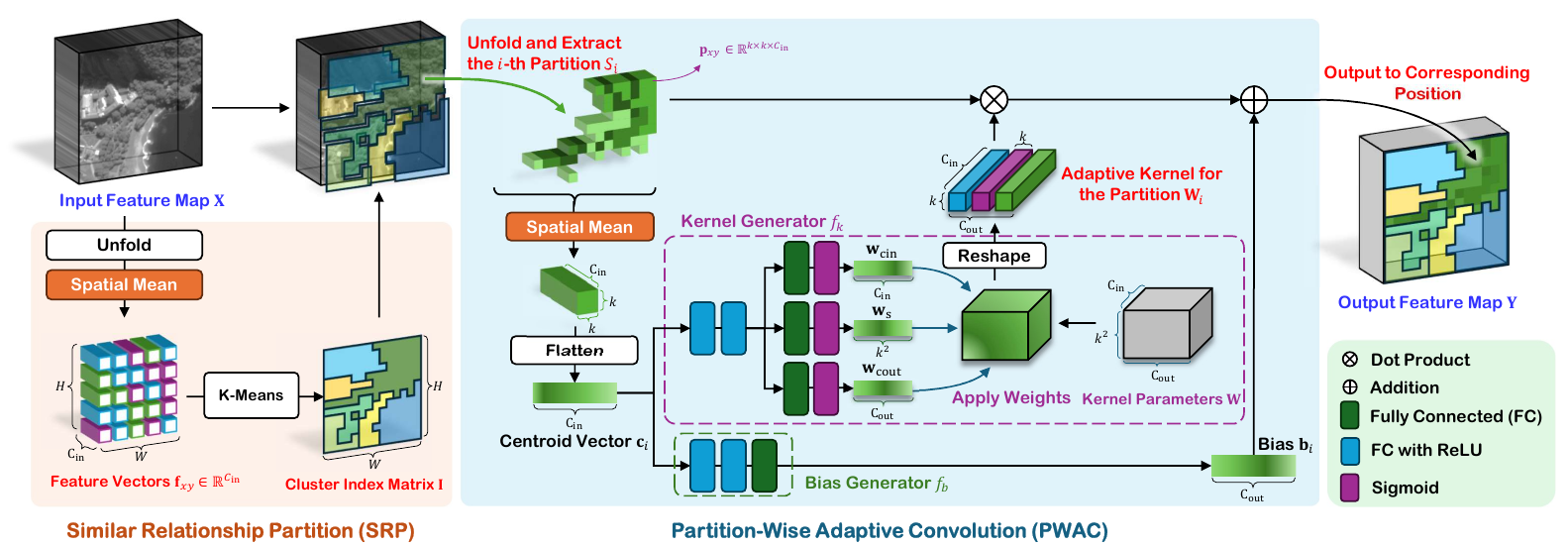}
  \caption{The overall workflow for a CANConv module. CANConv consists of two sub-modules: Similarity Relationship Partition (SRP) and Partition-Wise Adaptive Convolution (PWAC). In SRP, the input feature map is unfolded and reduced to obtain samples for clustering. PWAC is applied separately for each cluster distinguished by SRP. The figure demonstrates how PWAC adaptively generates convolution kernels and bias for a single cluster in the feature map.}
  \label{fig:CANConv}
  \vspace{-0.5cm}
\end{figure*}

This section introduces the design of the proposed CANConv module and CANNet. The former, CANConv, can be divided into two sub-modules: Similarity Relationship Partition (SRP) and Partition-Wise Adaptive Convolution (PWAC). For the latter, we construct CANNet by replacing standard convolution modules with CANConv, enabling the full utilization of non-local self-similarity in remote sensing pansharpening tasks. \cref{fig:CANConv} shows the overall process of the CANConv module.

\subsection{Content-Adaptive Non-Local Convolution}

\noindent{\bf Similarity Relationship Partition.} To effectively extract regional self-similarity relationships in remote sensing images, we first design the SRP module to cluster the input feature map. Let $\mathbf{X}\in\mathbb{R}^{H\times W\times C_\mathrm{in}}$ represent the input feature map, where $H$ and $W$ are the height and width of the feature map, and $C_\mathrm{in}$ is the number of input channels. Our goal is to compute a vector $\mathbf{f}_{xy}$ for each pixel in the feature map as the observations for clustering, where $(x,y)$ are the pixel coordinates (In subscripts, the comma and parenthesis are omitted to maintain conciseness). We define the unfold operation as extracting the neighborhood of each pixel with a $k\times k$ sliding window. To reduce the dimension of the neighborhood, a spatial mean pooling is performed on the neighborhoods to obtain $\mathbf{f}_{xy}$. The unfolding and pooling operation can be described as follows:
\begin{equation}
\mathbf{f}_{xy} = \frac{1}{k^2}\sum_{x'=-\lfloor k/2\rfloor}^{\lfloor k/2\rfloor}\sum_{y'=-\lfloor k/2\rfloor}^{\lfloor k/2\rfloor}\mathbf{X}_{x+x',y+y'},
\end{equation}
where the subscript of $\mathbf{X}$ indicates the $C_\mathrm{in}$-dimensional feature vector at the given position in the input feature map. Then we are able to cluster the obtained $H\times W$ observation vectors. K-Means algorithm~\cite{dingYinyangKmeansDropin2015} is chosen for clustering because of its simplicity and high efficiency. The results of clustering are represented by a cluster index matrix $\mathbf{I}\in \mathbb{N}^{H\times W}$, where the elements $\mathbf{I}_{xy}$ satisfy $0<\mathbf{I}_{xy}\leq K$, indicating the cluster number to which the pixel at $(x,y)$ belongs. Pixels with the same cluster index are in the same cluster, exhibiting non-local similarity. 

\begin{figure*}[t]
  \centering
  \includegraphics[width=1.0\linewidth]{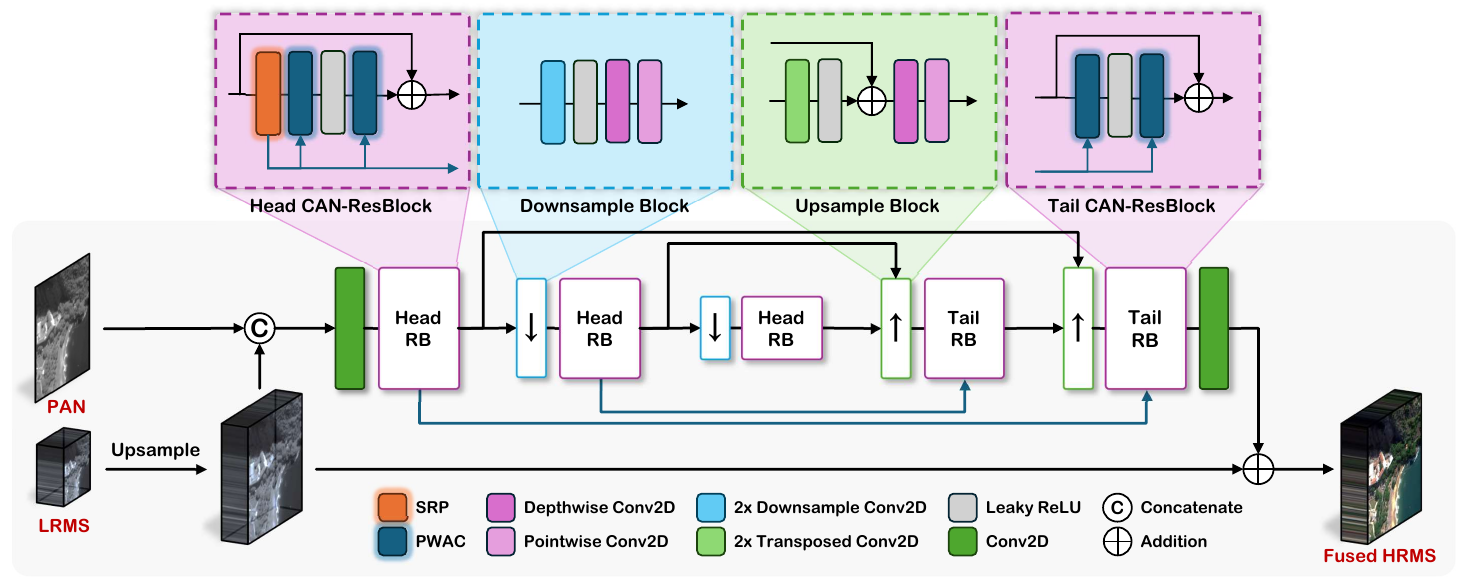}
  \caption{The overall architecture of CANNet. CANNet follows the classic U-Net design and features CAN-ResBlocks. Black arrows indicate the flow of feature maps, while blue arrows indicate the flow of cluster index matrices. The downsampling module halves the spatial resolution while doubling the number of channels, and the upsampling module does the opposite. Tail CAN-ResBlocks reuse cluster index matrices obtained in the Head CAN-ResBlock at the same level.}
  \label{fig:CANNet}
  \vspace{-0.5cm}
\end{figure*}

\noindent{\bf Partition-Wise Adaptive Convolution. } The PWAC sub-module first generates a set of convolution kernels for each self-similar partition based on its content, and then convolves all pixels within the partition using the generated kernel. During the convolution, we first unfold the feature map, and then group the neighborhoods by the cluster index matrix $\mathbf{I}$ obtained in the SRP sub-module. Mathematically, let $\mathbf{p}_{xy}\in\mathbb{R}^{k^2C_{\mathrm{in}}}$ denote the flat neighborhood of the input pixel at $(x,y)$, and pixel coordinates that belongs to the $i$-th cluster form a set
\begin{equation}
S_i=\{(x,y)|I_{xy}=i\},0<i\leq K.
\end{equation}

The overall process of PWAC can be represented as 
\begin{equation}
\mathbf{Y}_{xy}=\mathbf{p}_{xy}\otimes f(S_{\mathbf{I}_{xy}}),
\end{equation}
where $\mathbf{Y}_{xy}$ represents a $C_\mathrm{out}$-dimensional vector at coordinates $(x,y)$ in the output feature map,  $C_\mathrm{out}$ is the number of output channels, $f$ is a mapping that transforms the content in $S_i$ into a set of convolution kernels $\mathbf{W}_i\in\mathbb{R}^{C_{\mathrm{in}}k^2\times C_\mathrm{out}}$, and $\otimes$ denotes vector-matrix multiplication. To build $f$, we first compute the centroid vector of $S_i$ with
\begin{equation}
\mathbf{c}_i=\frac{1}{|S_i|}\sum_{(x,y)\in S_i}\mathbf{p}_{xy},
\label{eqn:centroid}
\end{equation}
where $|S_i|$ denotes the number of pixels belonging to the $i$-th partition. The centroid vector $\mathbf{c}_i\in\mathbb{R}^{k^2C_\mathrm{in}}$ can represent the content of the self-similar partition $S_i$. Then, we design a mapping $f_\mathrm{k}$ to adaptively generate the unique convolution kernel $\mathbf{W}_i\in\mathbb{R}^{k^2C_\mathrm{in}\times C_\mathrm{out}}$ for the $i$-th partition from $\mathbf{c}_i$. The details about $f_k$ will be discussed in the next part. Besides the convolution kernels, the biases $\mathbf{b_i}\in\mathbb{R}^{C_\mathrm{out}}$ of the convolution layer can also be adaptively generated. We directly utilize a multi-layer perceptron $f_\mathrm{b}$ to transform $\mathbf{c_i}$ into $\mathbf{b_i}$. Thus, the complete partition-wise adaptive convolution with bias can be represented as
\begin{equation}
\mathbf{Y}_{xy}=\mathbf{p}_{xy}\otimes f_\mathrm{k}(\mathbf{c}_{\mathbf{I}_{xy}})+f_\mathrm{b}(\mathbf{c}_{\mathbf{I}_{xy}}).
\label{eqn:ConvForward}
\end{equation}

As illustrated in the cluster visualization in \cref{sec:ClusterVisual}, there exist pixels with outlier features that fall into a few small clusters. Forcing the network to learn similarity information within such clusters may constrain its generalization ability. To address this issue, during training, the centroid vector of the cluster $\mathbf{c}_i$ is replaced with the global centroid in clusters where the number of pixels is less than a given threshold, as follows:
\begin{equation}
    \mathbf{c}_i=\frac{1}{HW}\sum_{0<x\leq H,0<y\leq W}\mathbf{p}_{xy}.
\label{eqn:global_centroid}
\end{equation}
\cref{eqn:global_centroid} is utilized to substitute \cref{eqn:centroid} if $|S_i|<\eta HW$, where $\eta$ represents the specified threshold ratio.

\noindent{\bf Lightweight Adaptive Kernel Generation. } In this paragraph we will introduce the design of the kernel generator $f_k$. It is possible to use a simple multi-layer perceptron with $k^2C_\mathrm{in}$ input features and $k^2C_\mathrm{in}C_\mathrm{out}$ output features to directly map $\mathbf{c}_i$ to $\mathbf{W}_i$. However, this approach includes too much amount of learnable parameters and makes the network harder to train. To reduce the parameter number, we utilize a global kernel parameter $\mathbf{W}\in\mathbb{R}^{C_{\mathrm{in}}\times k^2\times C_\mathrm{out}}$ and apply attention weight on it to obtain the unique convolution kernels $\mathbf{W}_i$ for each partition. We first use a multi-layer perceptron with three heads to generate three weight vectors corresponding to input channel dimension ($\mathbf{w}_\mathrm{cin}\in\mathbb{R}^{C_\mathrm{in}}$), spatial dimension ($\mathbf{w}_\mathrm{s}\in\mathbb{R}^{k^2}$), and output channel dimension ($\mathbf{w}_\mathrm{cout}\in\mathbb{R}^{C_\mathrm{out}}$). These three vectors are then used to weight the corresponding dimensions of the global convolution kernel parameters $\mathbf{W}$, and then the result is reshaped into $\mathbf{W}_i$. The weighting process can be formulated as
\begin{equation}
f_\mathrm{k}(\mathbf{c}_i)=\mathbf{W}_i=\mathbf{w}_\mathrm{cin}\circledast\mathbf{w}_\mathrm{s}\circledast\mathbf{w}_\mathrm{cout}\odot\mathbf{W},
\end{equation}
where $\circledast$ represents the Kronecker product, and $\odot$ represents element-wise multiplication.

\subsection{Network Architecture}
This section describes how to construct the network architecture of CANNet with non-local information utilization capabilities using the CANConv module. Similarity relationship partitioning and partition-wise adaptive convolution together form a complete CANConv module, which can directly replace standard convolution layers in existing convolutional networks. CAN-ResBlocks can replace the original ResBlocks~\cite{heDeepResidualLearning2016}. First, it extracts similarity relationships from input features with an SRP module, and then both PWACs use the same cluster index matrix. This design choice is made because the similarity relationships in CAN-ResBlock do not change significantly. Sharing the same index matrix helps save computational expenses. As illustrated in \cref{fig:CANNet}, CANNet follows the U-Net~\cite{ronnebergerUNetConvolutionalNetworks2015, wangSSconvExplicitSpectraltoSpatial2021} design, incorporating both skip connections for feature maps and skip connections for index matrices. PAN images and upsampled LRMS images are concatenated along the channel dimension to form the input feature map. Subsequently, through a series of down-sampling operations, CANConv can analyze and utilize similarity across different spatial scales and semantic levels. Then in upsampling, tail CAN-ResBlocks reuse cluster index matrices from downsampling to maintain and enhance the self-similar relationships. Finally, following the common detail-injection pattern in pansharpening CNNs\cite{hePansharpeningDetailInjection2019,dengDetailInjectionBasedDeep2021}, upsampled LRMS input is added to the output of the network to form the final HRMS image.

\section{Experiment}

\begin{table*}
\centering
\caption{Result benchmark on the WV3 dataset, evaluated with 20 reduced-resolution samples and 20 full-resolution samples. Best results are marked with bold font, and second-best results are marked with underline.}
\label{tab:wv3all}
\resizebox{0.8\linewidth}{!}{%
\begin{tblr}{
  row{2} = {c},
  cell{1}{1} = {r=2}{},
  cell{1}{2} = {c=3}{c},
  cell{1}{5} = {c=3}{c},
  cell{3}{2} = {c},
  cell{3}{3} = {c},
  cell{3}{4} = {c},
  cell{3}{5} = {c},
  cell{3}{6} = {c},
  cell{3}{7} = {c},
  cell{4}{2} = {c},
  cell{4}{3} = {c},
  cell{4}{4} = {c},
  cell{4}{5} = {c},
  cell{4}{6} = {c},
  cell{4}{7} = {c},
  cell{5}{2} = {c},
  cell{5}{3} = {c},
  cell{5}{4} = {c},
  cell{5}{5} = {c},
  cell{5}{6} = {c},
  cell{5}{7} = {c},
  cell{6}{2} = {c},
  cell{6}{3} = {c},
  cell{6}{4} = {c},
  cell{6}{5} = {c},
  cell{6}{6} = {c},
  cell{6}{7} = {c},
  cell{7}{2} = {c},
  cell{7}{3} = {c},
  cell{7}{4} = {c},
  cell{7}{5} = {c},
  cell{7}{6} = {c},
  cell{7}{7} = {c},
  cell{8}{2} = {c},
  cell{8}{3} = {c},
  cell{8}{4} = {c},
  cell{8}{5} = {c},
  cell{8}{6} = {c},
  cell{8}{7} = {c},
  cell{9}{2} = {c},
  cell{9}{3} = {c},
  cell{9}{4} = {c},
  cell{9}{5} = {c},
  cell{9}{6} = {c},
  cell{9}{7} = {c},
  cell{10}{2} = {c},
  cell{10}{3} = {c},
  cell{10}{4} = {c},
  cell{10}{5} = {c},
  cell{10}{6} = {c},
  cell{10}{7} = {c},
  cell{11}{2} = {c},
  cell{11}{3} = {c},
  cell{11}{4} = {c},
  cell{11}{5} = {c},
  cell{11}{6} = {c},
  cell{11}{7} = {c},
  cell{12}{2} = {c},
  cell{12}{3} = {c},
  cell{12}{4} = {c},
  cell{12}{5} = {c},
  cell{12}{6} = {c},
  cell{12}{7} = {c},
  cell{13}{2} = {c},
  cell{13}{3} = {c},
  cell{13}{4} = {c},
  cell{13}{5} = {c},
  cell{13}{6} = {c},
  cell{13}{7} = {c},
  cell{14}{2} = {c},
  cell{14}{3} = {c},
  cell{14}{4} = {c},
  cell{14}{5} = {c},
  cell{14}{6} = {c},
  cell{14}{7} = {c},
  cell{15}{2} = {c},
  cell{15}{3} = {c},
  cell{15}{4} = {c},
  cell{15}{5} = {c},
  cell{15}{6} = {c},
  cell{15}{7} = {c},
  cell{16}{2} = {c},
  cell{16}{3} = {c},
  cell{16}{4} = {c},
  cell{16}{5} = {c},
  cell{16}{6} = {c},
  cell{16}{7} = {c},
  cell{17}{2} = {c},
  cell{17}{3} = {c},
  cell{17}{4} = {c},
  cell{17}{5} = {c},
  cell{17}{6} = {c},
  cell{17}{7} = {c},
  % cell{18}{2} = {c},
  % cell{18}{3} = {c},
  % cell{18}{4} = {c},
  % cell{18}{5} = {c},
  % cell{18}{6} = {c},
  % cell{18}{7} = {c},
  % cell{19}{2} = {c},
  % cell{19}{3} = {c},
  % cell{19}{4} = {c},
  % cell{19}{5} = {c},
  % cell{19}{6} = {c},
  % cell{19}{7} = {c},
  % cell{20}{2} = {c},
  % cell{20}{3} = {c},
  % cell{20}{4} = {c},
  % cell{20}{5} = {c},
  % cell{20}{6} = {c},
  % cell{20}{7} = {c},
  hline{1,3,9,18-19} = {-}{}
}
\textbf{Methods}                                           & \textbf{Reduced-Resolution Metrics} &                            &                       & \textbf{Full-Resolution Metrics} &                        &                         \\
                                                           & \textbf{SAM}$\downarrow $           & \textbf{ERGAS}$\downarrow$ & \textbf{Q8}$\uparrow$ & $D_\lambda\downarrow$            & $D_s\downarrow$        & \textbf{HQNR}$\uparrow$ \\
EXP~\cite{EXP} & 5.800±1.881 & 7.155±1.878 & 0.627±0.092 & 0.0232±0.0066 & 0.0813±0.0318 & 0.897±0.036\\
% BT-H~\cite{aiazziMTFtailoredMultiscaleFusion2006}          & 4.920±1.425                         & 4.579±1.495                & 0.832±0.094           & 0.0574±0.0232                    & 0.0810±0.0374          & 0.867±0.054             \\
MTF-GLP-FS~\cite{vivoneFullScaleRegressionBased2018}       & 5.316±1.766                         & 4.700±1.597                & 0.833±0.092           & 0.0197±0.0078                    & 0.0630±0.0289          & 0.919±0.035             \\
TV~\cite{palssonNewPansharpeningAlgorithm2014}             & 5.692±1.808                         & 4.856±1.434                & 0.795±0.120           & 0.0234±0.0061                    & 0.0393±0.0227          & 0.938±0.027             \\
BDSD-PC~\cite{vivoneRobustBandDependentSpatialDetail2019}  & 5.429±1.823                         & 4.698±1.617                & 0.829±0.097           & 0.0625±0.0235                    & 0.0730±0.0356          & 0.870±0.053             \\
CVPR2019~\cite{fuVariationalPanSharpeningLocal2019} & 5.207±1.574& 5.484±1.505& 0.764±0.088 & 0.0297±0.0059& 0.0410±0.0136& 0.931±0.0183\\
% MTF-GLP-HPM-R~\cite{vivoneContrastErrorBasedFusion2014}    & 5.338±1.763                         & 5.230±3.016                & 0.835±0.092           & 0.0206±0.0082                    & 0.0630±0.0284          & 0.918±0.035             \\
LRTCFPan~\cite{LRTCFPan}                                   & 4.737±1.412                         & 4.315±1.442                & 0.846±0.091           & \uline{0.0176±0.0066}            & 0.0528±0.0258          & 0.931±0.031             \\
% BDPN~\cite{zhangPanSharpeningUsingEfficient2019}           & 4.206±0.858                         & 3.049±0.733                & 0.871±0.100           & 0.0364±0.0142                    & 0.0459±0.0192          & 0.920±0.031             \\
% MSDCNN~\cite{weiMultiscaleanddepthConvolutionalNeural2017} & 3.777±0.803                         & 2.761±0.688                & 0.890±0.090           & 0.0230±0.0091                    & 0.0467±0.0199          & 0.932±0.027             \\
PNN~\cite{masiPansharpeningConvolutionalNeural2016a}       & 3.680±0.763                         & 2.682±0.648                & 0.893±0.092           & 0.0213±0.0080                    & 0.0428±0.0147          & 0.937±0.021             \\
PanNet~\cite{yangPanNetDeepNetwork2017}                    & 3.616±0.766                         & 2.666±0.689                & 0.891±0.093           & \textbf{0.0165±0.0074}           & 0.0470±0.0213          & 0.937±0.027             \\
DiCNN~\cite{hePansharpeningDetailInjection2019}            & 3.593±0.762                         & 2.673±0.663                & 0.900±0.087           & 0.0362±0.0111                    & 0.0462±0.0175          & 0.920±0.026             \\
FusionNet~\cite{dengDetailInjectionBasedDeep2021}          & 3.325±0.698                         & 2.467±0.645                & 0.904±0.090           & 0.0239±0.0090                    & 0.0364±0.0137          & 0.941±0.020             \\
% MUCNN~\cite{wangSSconvExplicitSpectraltoSpatial2021}       & 3.206±0.681                         & 2.400±0.617                & 0.911±0.089           & 0.0258±0.0111                    & 0.0327±0.0140          & 0.942±0.021             \\
DCFNet~\cite{wuDynamicCrossFeature2021}                    & \uline{3.038±0.585}                 & \uline{2.165±0.499}        & 0.913±0.087           & 0.0187±0.0072                    & 0.0337±0.0054          & \uline{0.948±0.012}     \\
MMNet~\cite{mmnet} & 3.084±0.640 & 2.343±0.626 & \uline{0.916±0.086} & 0.0540±0.0232 &  \uline{0.0336±0.0115} & 0.914±0.028 \\
LAGConv~\cite{jinLAGConvLocalContextAdaptive2022}          & 3.104±0.559                         & 2.300±0.613                & 0.910±0.091           & 0.0368±0.0148                    & 0.0418±0.0152          & 0.923±0.025             \\
% PMAC~\cite{liangPMACNetParallelMultiscale2022}             & 3.073±0.623                         & 2.293±0.532                & 0.912±0.092           & 0.0540±0.0232                    & \uline{0.0336±0.0115}  & 0.914±0.028             \\
HMPNet~\cite{hmpnet}                                       & 3.063±0.577                         & 2.229±0.545                & 0.916±0.087   & 0.0184±0.0073                    & 0.0530±0.0055          & 0.930±0.011             \\
\textbf{Proposed}                                          & \textbf{2.930±0.593}                & \textbf{2.158±0.515}       & \textbf{0.920±0.084}  & 0.0196±0.0083                    & \textbf{0.0301±0.0074} & \textbf{0.951±0.013}    
\end{tblr}
}
\end{table*}

\subsection{Datasets, Metrics and Training Details}

To validate our approach, we construct datasets following Wald's protocol~\cite{waldFusionSatelliteImages1997, dengDetailInjectionBasedDeep2021} on data collected from the WorldView-3 (WV3), QuickBird (QB) and GaoFen-2 (GF2) satellites. Our datasets and data processing methods are downloaded from the PanCollection repository\footnote{\url{https://github.com/liangjiandeng/PanCollection}}~\cite{dengMachineLearningPansharpening2022}. Besides, we evaluated our method on three commonly used metrics in the field of pansharpening, including SAM~\cite{boardmanAutomatingSpectralUnmixing1993}, ERGAS~\cite{waldDataFusionDefinitions2002a} and Q4/Q8~\cite{garzelliHypercomplexQualityAssessment2009} for reduced-resolution dataset, and HQNR~\cite{arienzo2022full}, $D_s$, and $D_\lambda$ for full-resolution dataset. In addition, we utilize the $\ell_1$ loss function and Adam optimizer\cite{Adam} with a batch size of 32 in training. \textit{More details on datasets, metrics, and training can be found in supplementary materials \cref{sec:ExtraDetails}.}

\subsection{Results}

\begin{table}
\centering
\caption{Result benchmark on the QB dataset with 20 reduced-resolution samples. \textbf{Bold}: best, \uline{underline}: second best.}
\label{tab:qbreduced}
\setlength{\tabcolsep}{3pt}
\renewcommand\arraystretch{1.2}
\resizebox{\linewidth}{!}{%
\begin{tblr}{
  column{even} = {c},
  column{3} = {c},
  hline{1-2,8,17-18} = {-}{},
}
\textbf{Methods}                                  & \textbf{SAM}$\downarrow$ & \textbf{ERGAS}$\downarrow$ & \textbf{Q4}$\uparrow$ \\
EXP~\cite{EXP}                                               & 8.435±1.925              & 11.819±1.905               & 0.584±0.075           \\
% MTF-GLP-HPM                                       & 8.364±1.994              & 8.313±0.971                & 0.795±0.084           \\
% BT-H~\cite{aiazziMTFtailoredMultiscaleFusion2006}                                              & 7.194±1.552              & 7.401±0.838                & 0.833±0.088           \\
TV~\cite{palssonNewPansharpeningAlgorithm2014} & 7.565±1.535& 7.781±0.699& 0.820±0.090\\
MTF-GLP-FS~\cite{vivoneFullScaleRegressionBased2018}                                        & 7.793±1.816              & 7.374±0.724                & 0.835±0.088           \\
BDSD-PC~\cite{vivoneRobustBandDependentSpatialDetail2019}                                           & 8.089±1.980              & 7.515±0.800                & 0.831±0.090           \\
CVPR19~\cite{fuVariationalPanSharpeningLocal2019} & 7.998±1.820              & 9.359±1.268                & 0.737±0.087           \\
LRTCFPan~\cite{LRTCFPan}                                          & 7.187±1.711              & 6.928±0.812                & 0.855±0.087           \\
PNN~\cite{masiPansharpeningConvolutionalNeural2016a}                                               & 5.205±0.963              & 4.472±0.373                & 0.918±0.094           \\
PanNet~\cite{yangPanNetDeepNetwork2017}                                            & 5.791±1.184              & 5.863±0.888                & 0.885±0.092           \\
DiCNN~\cite{hePansharpeningDetailInjection2019}                                             & 5.380±1.027              & 5.135±0.488                & 0.904±0.094           \\
FusionNet~\cite{dengDetailInjectionBasedDeep2021}                                         & 4.923±0.908              & 4.159±0.321                & 0.925±0.090           \\
DCFNet~\cite{wuDynamicCrossFeature2021}                                            & \uline{4.512±0.773}      & 3.809±0.336                & 0.934±0.087           \\
MMNet~\cite{mmnet} & 4.557±0.729 & 3.667±0.304 & 0.934±0.094 \\
LAGConv~\cite{jinLAGConvLocalContextAdaptive2022}                                           & 4.547±0.830              & 3.826±0.420                & 0.934±0.088           \\
HMPNet~\cite{hmpnet}                                            & 4.617±0.404              & \textbf{3.404±0.478}       & \uline{0.936±0.102}   \\
\textbf{Proposed}                                 & \textbf{4.507±0.835}     & \uline{3.652±0.327}        & \textbf{0.937±0.083}  
\end{tblr}
}
\end{table}

\begin{table}
\centering
\caption{Result benchmark on the GF2 dataset with 20 reduced-resolution samples. \textbf{Bold}: best, \uline{underline}: second best.}
\label{tab:gf2reduced}
\setlength{\tabcolsep}{3pt}
\renewcommand\arraystretch{1.2}
\resizebox{\linewidth}{!}{%
\begin{tblr}{
  column{even} = {c},
  column{3} = {c},
  hline{1-2,8,17-18} = {-}{},
  % hline{1-2,8,14-15} = {-}{}
}
\textbf{Method}   & \textbf{SAM↓}        & \textbf{ERGAS↓}      & \textbf{Q4↑}         \\
EXP~\cite{EXP}               & 1.820±0.403          & 2.366±0.554          & 0.812±0.051          \\
% BT-H~\cite{aiazziMTFtailoredMultiscaleFusion2006}              & 1.649±0.360          & 1.528±0.409          & 0.918±0.025          \\
TV~\cite{palssonNewPansharpeningAlgorithm2014}                & 1.918±0.398          & 1.745±0.405          & 0.905±0.027\\
% MTF-GLP-HPM       & 1.873±0.405          & 1.787±0.375          & 0.876±0.033          \\
MTF-GLP-FS~\cite{vivoneFullScaleRegressionBased2018}        & 1.655±0.385          & 1.589±0.395          & 0.897±0.035          \\
BDSD-PC~\cite{vivoneRobustBandDependentSpatialDetail2019}           & 1.681±0.360          & 1.667±0.445          & 0.892±0.035          \\
CVPR19~\cite{fuVariationalPanSharpeningLocal2019}            & 1.598±0.353          & 1.877±0.448          & 0.886±0.028          \\
LRTCFPan~\cite{LRTCFPan}          & 1.315±0.283          & 1.301±0.313          & 0.932±0.033          \\
PNN~\cite{masiPansharpeningConvolutionalNeural2016a}               & 1.048±0.226          & 1.057±0.236          & 0.960±0.010          \\
PanNet~\cite{yangPanNetDeepNetwork2017}            & 0.997±0.212          & 0.919±0.191          & 0.967±0.010          \\
DiCNN~\cite{hePansharpeningDetailInjection2019}             & 1.053±0.231          & 1.081±0.254          & 0.959±0.010          \\
FusionNet~\cite{dengDetailInjectionBasedDeep2021}         & 0.974±0.212          & 0.988±0.222          & 0.964±0.009          \\
DCFNet~\cite{wuDynamicCrossFeature2021}            & 0.872±0.169          & 0.784±0.146          & 0.974±0.009          \\
MMNet~\cite{mmnet} & 0.993±0.141 & 0.812±0.119 & 0.969±0.020 \\
LAGConv~\cite{jinLAGConvLocalContextAdaptive2022}           & \uline{0.786±0.148}  & 0.687±0.113          & 0.980±0.009          \\
HMPNet~\cite{hmpnet}            &0.803±0.141           &\textbf{0.564±0.099}  & \uline{0.981±0.030}  \\
\textbf{Proposed} & \textbf{0.707±0.148} & \uline{0.630±0.128}  & \textbf{0.983±0.006} 
\end{tblr}
}
\end{table}

\begin{figure*}
  \centering
  \includegraphics[width=1\linewidth]{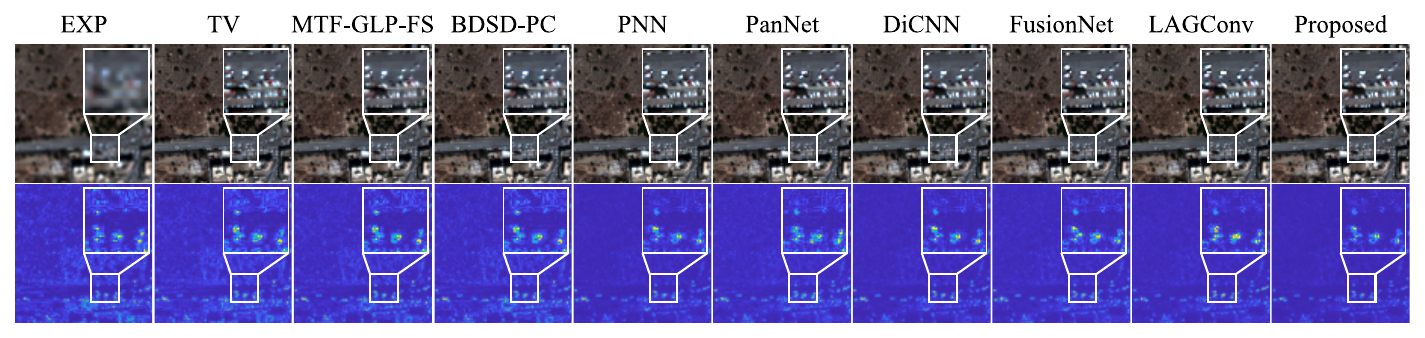}
  \caption{Qualitative result comparison between representative methods on the WV3 reduced-resolution dataset. The first row presents RGB outputs, while the second row shows the residual compared to the ground truth. Refer to supplementary material for more comparison. }
  \label{fig:WV3VisualCompare}
  % \vspace{-0.5cm}
\end{figure*}

The performance of the proposed CANNet method is showcased through extensive evaluations on three benchmark datasets: WV3, QB, and GF2. \cref{tab:wv3all,tab:qbreduced,tab:gf2reduced} present a comprehensive comparison of CANNet with various state-of-the-art methods, including both traditional and deep learning approaches. These results confirm the robustness of CANNet in handling different datasets and its consistent ability to produce high-quality pan-sharpened images. Furthermore, visual comparisons in \cref{fig:WV3VisualCompare} illustrate that CANNet generates results closer to the ground truth. This is a testament to CANNet's utilization of non-local self-similarity information, endowing it with outstanding spatial fidelity capabilities. \textit{For more benchmarks and visualizations, please refer to the supplementary material \cref{sec:AdditionalResults}.}

\subsection{Discussions}

\begin{figure}
   \centering
   \begin{subfigure}{0.3\linewidth}
       \centering
       \includegraphics[width=1\linewidth]{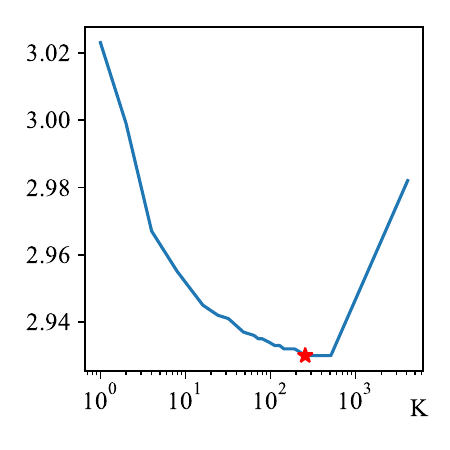}
       \caption{SAM}
       \label{fig:SamByK}
   \end{subfigure}
    \begin{subfigure}{0.3\linewidth}
       \centering
       \includegraphics[width=1\linewidth]{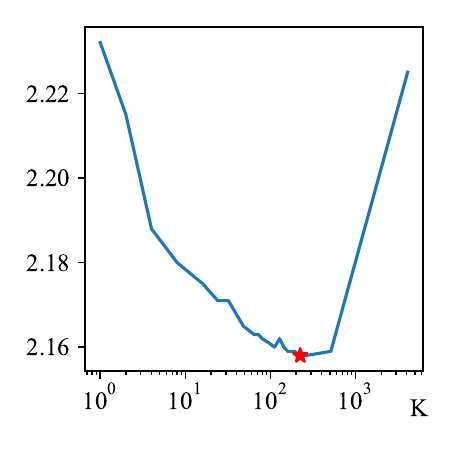}
       \caption{ERGAS}
       \label{fig:ErgasByK}
   \end{subfigure}
   \begin{subfigure}{0.3\linewidth}
       \centering
       \includegraphics[width=1\linewidth]{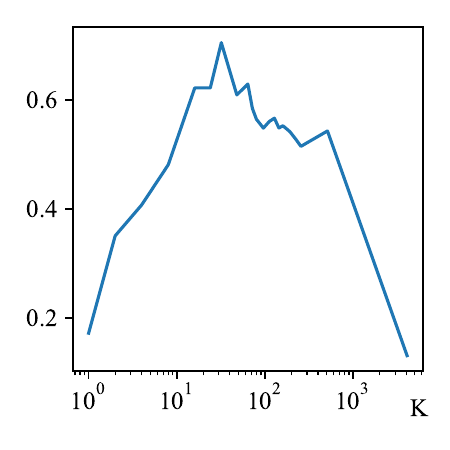}
       \caption{Inference Time}
       \label{fig:TimeByK}
   \end{subfigure}
   \caption{Variations of SAM, ERGAS and inference time on the WV3 reduced-resolution dataset with changing cluster number $K$. The optimal metrics are obtained around $K=256$. }
   \label{fig:MetricByK}
\end{figure}

\begin{table}
\centering
\caption{Result of replacing convolution modules in other backbones with CANConv on WV3 reduced-resolution dataset.}
\label{tab:wv3replace}
\resizebox{\linewidth}{!}{%
\begin{tblr}{
  column{even} = {c},
  column{3} = {c},
  hline{1-2,4,6,8} = {-}{},
}
\textbf{Method} & \textbf{SAM↓} & \textbf{ERGAS↓} & \textbf{Q8↑} \\
DiCNN           & 3.593±0.762   & 2.673±0.663     & 0.900±0.087  \\
CAN-DiCNN       & 3.371±0.694   & 2.492±0.609     & 0.906±0.086  \\
FusionNet       & 3.325±0.698   & 2.467±0.645     & 0.904±0.090  \\
CAN-FusionNet   & 3.293±0.703   & 2.407±0.591     & 0.907±0.086  \\
LAGConv         & 3.104±0.559   & 2.300±0.613     & 0.910±0.091  \\
CAN-LAGConv     & 3.051±0.617   & 2.251±0.564     & 0.914±0.086  
\end{tblr}
}
\end{table}

\noindent\textbf{Cluster Count of CANNet}: We explored the effectiveness of CANNet by varying the number of clusters $K$ in the K-Means algorithm. We trained the network with $K=32$ and then inferred on the test set with different values of $K$. In \cref{fig:SamByK,fig:ErgasByK}, we found that performance improves as $K$ increases when small, but deteriorates when too large. Small $K$ yields results close to no clustering, as dissimilar pixels are grouped together, hindering effective kernel adaptation. Excessively large $K$ results in too few pixels per cluster, impeding information gathering. The leftmost and rightmost points on the curves correspond to the cases of using a global convolutional kernel and a fully spatial adaptive convolutional kernel, respectively. The experiments show the effectiveness of clustering in transferring information between similar pixels. It is worth noting that the optimal value of $K$ on the test set is about four times that of the training set. This is because the images in the test set cover a larger spatial range and more diverse features.

\noindent\textbf{Analysis on Complexity}: Compared to other adaptive convolutions, additional computation complexity of CANConv mainly attributes to K-Means, which has a theoretical complexity of $O(tKnd)$, where $t$, $K$ and $n$ are the number of iterations, clusters and samples, and $d$ is the dimension of samples. The adaption from \cite{dingYinyangKmeansDropin2015} is used to reduce $t$ by observing the change in cluster centers. \cref{fig:TimeByK} presents the observed inference time on an RTX3090 with changing $K$. It is noticeable that the time decreases when $K$ is high, because the decline in $t$ counteracts the increase in $K$.
 
\noindent\textbf{Replacing Standard Convolution}: To verify the performance of CANConv as a standalone convolution module, we try to replace convolution modules in various pansharpening backbones with CANConv. The results are presented in \cref{tab:wv3replace}. For DiCNN~\cite{hePansharpeningDetailInjection2019}, a three-layer convolutional neural network, replacing one convolutional layer with CANConv significantly improves network performance. In FusionNet~\cite{dengDetailInjectionBasedDeep2021}, which consists of four standard ResBlocks, replacing some with CAN-ResBlocks also yields noticeable improvements. Even for LAGNet~\cite{jinLAGConvLocalContextAdaptive2022}, which already incorporates spatial adaptive convolution, substituting some spatial adaptive ResBlocks with CAN-ResBlocks using CANConv further enhances performance by leveraging additional self-similarity information. \textit{The supplementary material \cref{sec:ExtraReplace} contains details of this experiment.}

\begin{figure*}
  \centering
  \includegraphics[width=1\linewidth]{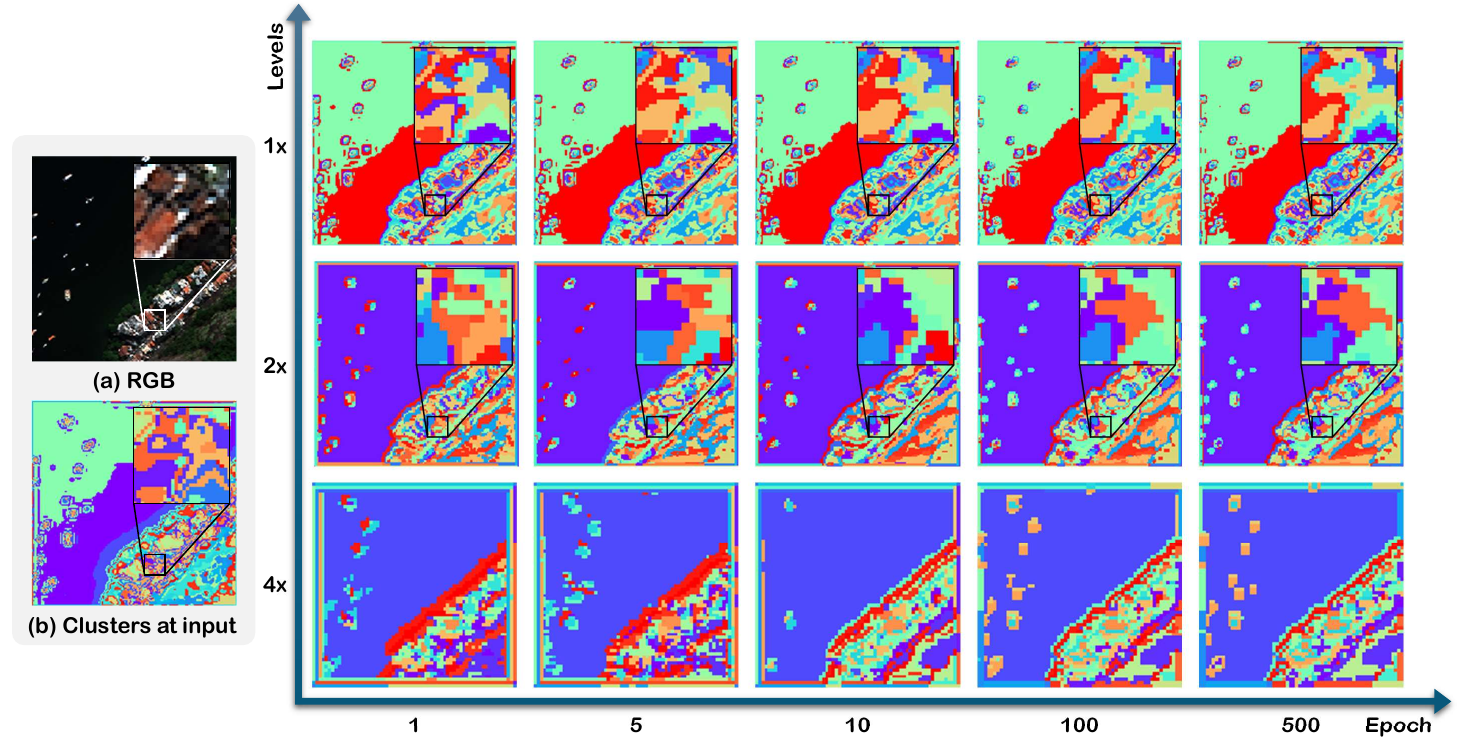}
  \caption{Visual representations of cluster index matrices in CANNet at different downsample levels and training epochs. (a) RGB appearance for the sample image from the WV3 reduced-resolution test dataset. (b) Clustering results on concatenated raw PAN and LRMS input images without being transformed by convolution layers. The color indicates the cluster to which the pixel belongs.  }
  \label{fig:ClustersVisual}
  \vspace{-0.2cm}
\end{figure*}

\label{sec:ClusterVisual}

\noindent\textbf{Cluster Visualization}: \cref{fig:ClustersVisual} displays the similarity partition distribution by the SRP sub-module at various CANNet levels, suggesting CANConv effectively classifies different region types on the feature map. Moreover, considering the edge morphology of the regions, shallow layers emphasize color and texture similarity, whereas in the deeper layers the clustering results encompass more semantic information. The cluster index matrix stabilizes and incorporates more semantics during network training. Although most pixels form large, similar clusters, some outliers fall into smaller, less similar clusters. Border clusters exhibit straight-line shapes due to zero padding when unfolding.

\begin{table}
\centering
\caption{Ablation experiment on WV3 reduced-resolution dataset.}
\label{tab:ablation1}
\resizebox{\linewidth}{!}{%
\begin{tblr}{
  column{even} = {c},
  column{3} = {c},
  hline{1-2,5} = {-}{},
}
\textbf{Ablation}  & \textbf{SAM↓} & \textbf{ERGAS↓} & \textbf{Q8↑} \\
None               & 2.930±0.593   & 2.158±0.515     & 0.920±0.084  \\
No SRP, $K=1$      & 3.023±0.606   & 2.232±0.567     & 0.919±0.083  \\
No SRP, $K=\infty$ & 2.982±0.608   & 2.225±0.546     & 0.918±0.084  
\end{tblr}
}
\end{table}

\begin{table}
\centering
\caption{Ablation experiment on WV3 full-resolution dataset.}
\resizebox{\linewidth}{!}{%
\begin{tblr}{
  column{2-4} = {c},
  cell{3}{2} = {c=3}{},
  hline{1-2,5} = {-}{},
}
\textbf{Ablation}                    & $D_l\downarrow$         & $D_s\downarrow$         & \textbf{HQNR↑}          \\
None                        & 0.0196±0.0083 & 0.0301±0.0074 & 0.951±0.013   \\
Replace $f_k$~with MLP      & Not Converged &               &               \\
$\eta=0$ & 0.0169±0.0064 & 0.0400±0.0053 & 0.944±0.009
\end{tblr}
}
\label{tab:ablation2}
\end{table}

\noindent\textbf{Ablation Study}: We ablated different parts in our method to prove their effectiveness, and the results are shown in \cref{tab:ablation1,tab:ablation2}. To ablate the SRP sub-module, the results in \cref{fig:MetricByK,tab:ablation1} reveal cases where SRP is disabled, including treating the whole input feature map as a single cluster ($K=1$, leftmost) or considering each pixel as a separate cluster ($K=\infty$, rightmost). The network's performance experiences a sharp decline when SRP is disabled, underscoring the effectiveness of the SRP. Regarding the ablation of the PWAC process, we attempted to replace $f_k$ with a direct multi-layer perceptron comprising the same number of layers and inner features. However, this replacement increased the learnable parameters to about $10\times$, causing the network fail to converge. To demonstrate the impact of replacing centroid vectors in small clusters, we sought to mitigate this behavior by setting $\eta = 0$ in training.

\section{Conclusion}

In conclusion, our paper introduces CANConv, a novel non-local adaptive convolution module that addresses the limitations of conventional spatial adaptive convolution methods by simultaneously incorporating spatial adaptability and non-local self-similarity. Integrated into the CANNet network, CANConv leverages multi-scale self-similarity information, offering a comprehensive solution for the remote sensing pansharpening task. Our contributions include the analysis of non-local self-similarity relationships in remote sensing images, validating the CANConv method on multiple datasets, and demonstrating its state-of-the-art performance compared to existing pansharpening methods. 

% \subsection*{Acknowledgement}

\noindent\textbf{Acknowledgement}: This work is supported by NSFC (12271083).

% \clearpage
\FloatBarrier
% \clearpage
{
    \small
    \bibliographystyle{ieeenat_fullname}
    \bibliography{main}
}

\clearpage

\clearpage
\maketitlesupplementary

\begin{abstract}
    The supplementary materials offer further insights into the CANConv method proposed in our paper. We delve into a thorough analysis of the clustering model within CANConv, offering additional details on experimental settings, encompassing datasets and training parameters. Furthermore, we introduce alternative methods used in benchmarking results and elaborate on the settings for discussion experiments. Lastly, we provide additional benchmarks on QB and GF2 full-resolution datasets and visual comparisons of results among benchmarked methods.
\end{abstract}

\section{Analysis on KNN and K-Means}

Many previous works\cite{plotzNeuralNearestNeighbors2018, zhouCrossScaleInternalGraph2020, liCrossPatchGraphConvolutional2021} have used the K-Nearest Neighbors (KNN) model to capture similarity relationships in feature maps, while our method employs the K-Means clustering algorithm, which has significant differences between the two approaches.

\noindent\textbf{K-Nearest Neighbors (KNN)}: In traditional machine learning, the KNN algorithm is commonly used for classification and regression tasks by determining the classification or regression value based on the values of the k-nearest neighbors to the sample to be predicted. Previous graph convolution methods used the KNN model to model similarity relationships between patches in images, requiring the computation of pairwise distances between all patches and finding the $k$ most similar 
patches for each patch, incurring a large computational cost. These methods achieve information propagation through convolution layers by concatenating patches along the channel dimension, increasing the spatial dimensions of the feature map by a factor of $k$, and pre-trained weights cannot adapt to changes in $k$.

\noindent\textbf{K-Means}: This paper adopts a clustering approach to model similarity relationships between patches and selects the simple unsupervised K-Means algorithm to perform clustering, dividing samples into $K$ clusters to maximize similarity within each cluster and minimize similarity between clusters. The typical usage of the K-Means algorithm in traditional machine learning involves iteratively computing $K$ cluster centers on the training set and directly finding the cluster center closest to the sample during prediction. In deep learning for vision tasks, the dataset contains a vast number of patches, and the data distribution of the feature map changes with training epochs. To achieve maximum flexibility to adapt to different input data, \textit{we choose to cluster all patches in a single image during both training and inference, recomputing cluster centers, rather than only comparing samples with those in the training set during inference.} As an unsupervised clustering algorithm, K-Means does not guarantee that the same content will be assigned the same cluster number in each image. For example, the ocean on image 1 may belong to cluster 3, while the ocean on image 2 may belong to cluster 6. This limitation prevents us from specifying convolution kernels based on cluster numbers in the PWAC module; instead, we generate convolution kernels adaptively based on the content of the clusters. The benefit of this approach is the \textit{decoupling of learnable parameters from the value of $K$}. We don't need to store $K$ sets of convolution kernel parameters, using only one set of parameters to generate different convolution kernels for all clusters, and allowing for changing the value of $K$ at any time to adapt to varying inputs.

\section{Backpropagation in Cluster Algorithm}

Since K-Means is an unsupervised clustering algorithm, we have to carefully handle the gradients. Though K-Means is not differentiable, it will not block the backpropagation in the network, since its output $\mathbf{I}$ is only used for index-selecting $\mathbf{X}$ to get $\mathbf{c}_i$. It is still possible to estimate gradients of $\mathbf{X}$ directly from $\mathbf{c}_i$, while ignoring gradients from $\mathbf{I}$. The whole process can be written as 

\noindent\begin{equation*}
\begin{aligned}
    % \mathbf{I}&=\operatorname{SRP}(\mathbf{X}),\mathbf{Y}=\operatorname{PWAC}(\mathbf{X},\mathbf{I}),\\
    \frac{\partial\mathcal{L}}{\partial\mathbf{X}}&=\frac{\partial\mathcal{L}}{\partial\mathbf{c}_i}\bigg(\frac{\partial\mathbf{c}_i}{\partial\mathbf{X}}+\underbrace{\frac{\partial\mathbf{c}_i}{\partial\mathbf{I}}\cdot\frac{\partial\mathbf{I}}{\partial\mathbf{X}}}_{\mathrm{Ignored}}\bigg).
\end{aligned}
\end{equation*}

\vspace{-0.2cm}

In practice, we compute gradients of $\mathbf{c}_i$ and $\mathbf{p}_{xy}$ in $\mathbf{X}$ using the following formulas:

\begin{equation}
\begin{aligned}
    \frac{\partial \mathcal{L}}{\partial\mathbf{c}_{i}}=&\left(\sum_{(x,y)\in S_i}\mathbf{p}_{xy}^\top\times\frac{\partial \mathcal{L}}{\partial\mathbf{Y}_{xy}}\right)\frac{\partial\mathbf{W}_i}{\partial\mathbf{c}_i}\\
    &+\left(\sum_{(x,y)\in S_i}\mathbf{p}_{xy}\right)\frac{\partial\mathbf{b}_i}{\partial\mathbf{c}_i},
\end{aligned}
\end{equation}

\begin{equation}
\frac{\partial \mathcal{L}}{\partial\mathbf{p}_{xy}}=\frac{\partial \mathcal{L}}{\partial\mathbf{Y}_{xy}}\times\mathbf{W}_{\mathbf{I}_{xy}}^\top+\frac{1}{|S_{\mathbf{I}_{xy}}|}\frac{\partial \mathcal{L}}{\partial\mathbf{c}_{\mathbf{I}_{xy}}},
\end{equation}

where $\mathcal{L}$ refers to the loss function. $\partial\mathbf{W}_i/\partial\mathbf{c}_i$ , $\partial\mathbf{W}_i/\partial\mathbf{c}_i$ and the gradients of the learnable parameters in $f_k$, $f_b$ and $\mathbf{W}$, can be easily calculated using automatic differentiation frameworks. Experimental results indicate that ignoring gradients related to clustering and index operations does not affect the convergence of the network.

\section{Details on Experiments and Discussion}

\label{sec:ExtraDetails}

\begin{table*}
\centering
\caption{Introduction for pansharpening methods involved in the benchmark.}
\label{tab:methods}
\begin{tblr}{
  colspec = {X[2.8,l] X[1.5,c] X[1,j,c] X[10,j,m]},
  hline{1-2,16} = {-}{},
}
\textbf{ Method } & \textbf{ Category } & \textbf{ Year } & \textbf{ Introduction }                                                                                                                                      \\
EXP~\cite{EXP}               &                     & 2002            & Simply upsamples the MS image.                                                                                                                               \\
MTF-GLP-FS~\cite{vivoneFullScaleRegressionBased2018}        & MRA                 & 2018            & Estimates the injection coefficients at full resolution rather than reduced resolution.                                                                      \\
TV~\cite{palssonNewPansharpeningAlgorithm2014}                & VO                  & 2013            & Employs total variation as a regularization technique for addressing an ill-posed problem defined by a commonly utilized explicit model for image formation. \\
BDSD-PC~\cite{vivoneRobustBandDependentSpatialDetail2019}           & CS                  & 2018            & Addresses the limitations of the band-dependent spatial-detail (BDSD) method in images with more than four spectral bands.                                   \\
CVPR2019~\cite{fuVariationalPanSharpeningLocal2019}          & VO                  & 2019            & Integrates a more precise spatial preservation strategy by considering local gradient constraints within distinct local patches and bands.                   \\
LRTCFPan~\cite{LRTCFPan}          & VO                    & 2023            & Utilizes low-rank tensor completion (LRTC) as the foundation and incorporating various regularizers for enhanced performance.                                \\
PNN~\cite{masiPansharpeningConvolutionalNeural2016a}               & ML                  & 2016            & The first convolutional neural network (CNN) for pansharpening with three convolutional layers.                                                              \\
PanNet~\cite{yangPanNetDeepNetwork2017}            & ML                  & 2017            & Deeper CNN for pansharpening.                                                                                                                                \\
DiCNN~\cite{hePansharpeningDetailInjection2019}             & ML                  & 2019            & Introduces the detail injection procedure into pansharpening CNNs.                                                                                           \\
FusionNet~\cite{dengDetailInjectionBasedDeep2021}         & ML                  & 2021            & Combines ML techniques with traditional fusion schemes like CS and MRA.                                                                                      \\
DCFNet~\cite{wuDynamicCrossFeature2021}            & ML                  & 2021            & Considers the connections of information between high-level semantics and low-level features through the incorporation of multiple parallel branches.        \\
MMNet~\cite{mmnet}             & ML                  & 2022            & A model-driven deep unfolding network with memory-augmentation.                                                                                              \\
LAGConv~\cite{jinLAGConvLocalContextAdaptive2022}           & ML                  & 2022            & Adaptive convolution with enhanced ability to leverage local information and preserve global harmony.                                                        \\
HMPNet~\cite{hmpnet}           & ML                  & 2023            & An interpretable model-driven deep network tailored for the fusion of hyperspectral (HS), multispectral (MS), and panchromatic (PAN) images   
\end{tblr}
\end{table*}

\subsection{Datasets}  

We conducted experiments on data collected from the WorldView-3 (WV3), QuickBird (QB) and GaoFen-2 (GF2) satellites. The datasets consist of images cropped from entire remote sensing images, divided into training and testing sets. The training set comprises PAN/LRMS/GT image pairs obtained by downsampling simulation, with dimensions of $64\times 64$, $16\times 16\times C$ and $64\times 64\times C$, respectively. The WV3 training set contains approximately 10,000 pairs of eight-channel images ($C=8$), while the QB training set contains around 17,000 pairs of four-channel images ($C=4$). GF2 training set has about 20,000 pairs of four-channel images ($C=4$). The reduced-resolution testing set for each satellite consists of 20 downsampling simulated PAN/LRMS/GT image pairs with various representative land covers, with dimensions of $256\times 256$, $64\times 64\times C$, and $256\times 256\times C$, respectively. The full-resolution test set includes 20 pairs of original PAN/LRMS images with dimensions of $512\times 512$ and $128\times 128$. Our datasets and data processing methods are downloaded from the PanCollection repository~\cite{dengMachineLearningPansharpening2022}.

\subsection{Training Details}

When training CANNet on the WV3 dataset, we utilized the $\ell_1$ loss function and Adam optimizer~\cite{Adam} with a batch size of 32. The initial learning rate was set at $10^{-3}$, which was reduced to $10^{-4}$ after 250 epochs. The total duration of the training was 500 epochs. Regarding the network architecture, we set the number of channels in the hidden layers to 32, the number of clusters $K$ during training was set to 32 and the threshold $\eta$ was 0.005. To encourage stable clustering learning, we recalculated and updated the cluster indices every 10 epochs during training. For the QB dataset, we maintained a constant learning rate of $5\times 10^{-4}$ and only trained for 200 epochs, while all other parameters were kept the same as in the WV3 dataset.

Here we provide details regarding performing K-Means in CANConv. 
We select initial cluster centers using the K-Means++\cite{kmeanspp} method. We initialize the centers separately for different samples at different layers, because they capture distinct features (As show in \cref{fig:ClustersVisual}). We stop iterating when less than 1\% of cluster assignments are changed. In practice, it typically takes 20-25 iterations to converge.

% \vspace{-0.2cm}

\subsection{Compared Methods}

% \vspace{-0.2cm}

\cref{tab:methods} provides a brief overview of pansharpening methods compared in the main text. We compare the proposed CANNet with both traditional and machine learning (ML) methods. We choose representative traditional methods from three categories including CS, VO and MRA. We also select classic and recent ML methods for benchmarking.

% \vspace{-0.2cm}

\subsection{Replacing Standard Convolution}

\label{sec:ExtraReplace}

% \vspace{-0.2cm}

\begin{figure}
   \centering
   \begin{subfigure}{0.49\linewidth}
       \centering
       \includegraphics[width=1\linewidth]{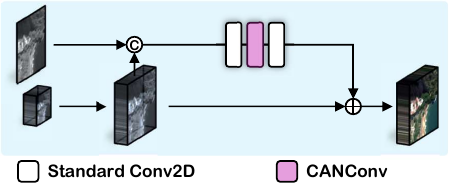}
       \caption{CAN-DiCNN}
   \end{subfigure}
   \begin{subfigure}{0.49\linewidth}
       \centering
       \includegraphics[width=1\linewidth]{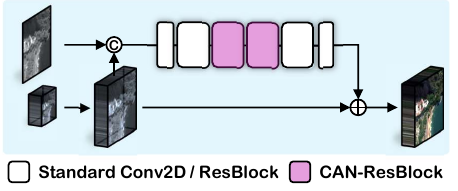}
       \caption{CAN-FusionNet}
   \end{subfigure}
   \begin{subfigure}{0.49\linewidth}
       \centering
       \includegraphics[width=1\linewidth]{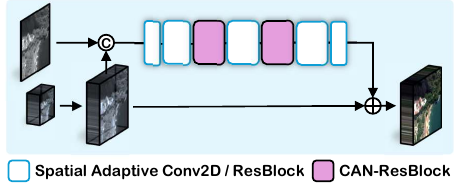}
       \caption{CAN-LAGNet}
   \end{subfigure}

   \caption{Replacing standard convolution module with CANConv to leverage non-local self-similarity information. Highlighted modules are replaced with CANConv or CAN-ResBlock in the experiment.}
   \label{fig:Replace}
   % \vspace{-0.4cm}
\end{figure}

This section presents the details of the discussion experiment on replacing standard convolution. \cref{fig:Replace} shows which layers or blocks are replaced with their CANConv counterparts. In hyperparameter tuning, we increased the learning rate on CAN-DiCNN to $10^{-3}$ to foster faster convergence, and kept all other parameters the same as the original network. The number of clusters $K$ was set to 32 and the threshold $\eta$ was 0.005.

\begin{table}
    \centering
    \caption{Result benchmark on the QB dataset with 20 full-resolution samples. \textbf{Bold}: best, \uline{underline}: second best.}
    \label{tab:qbfull}
    \resizebox{\linewidth}{!}{%
        \begin{tblr}{
            column{even} = {c},
            column{3} = {c},
            hline{1-2,8,16-17} = {-}{},
                }
            \textbf{Method}                                           & $D_\lambda\downarrow$  & $D_s\downarrow$        & \textbf{HQNR}$\uparrow$ \\
            EXP~\cite{EXP}                                            & 0.0436±0.0089          & 0.1502±0.0167          & 0.813±0.020             \\
            TV~\cite{palssonNewPansharpeningAlgorithm2014}            & 0.0465±0.0146          & 0.1500±0.0238          & 0.811±0.034             \\
            MTF-GLP-FS~\cite{vivoneFullScaleRegressionBased2018}      & 0.0550±0.0142          & 0.1009±0.0265          & 0.850±0.037             \\
            BDSD-PC~\cite{vivoneRobustBandDependentSpatialDetail2019} & 0.1975±0.0334          & 0.1636±0.0483          & 0.672±0.058             \\
            CVPR19~\cite{fuVariationalPanSharpeningLocal2019}         & 0.0498±0.0119          & 0.0783±0.0170          & 0.876±0.023             \\
            LRTCFPan~\cite{LRTCFPan}                                  & \textbf{0.0226±0.0117} & 0.0705±0.0351          & \uline{0.909±0.044}     \\
            PNN~\cite{masiPansharpeningConvolutionalNeural2016a}      & 0.0577±0.0110          & 0.0624±0.0239          & 0.884±0.030             \\
            PanNet~\cite{yangPanNetDeepNetwork2017}                   & 0.0426±0.0112          & 0.1137±0.0323          & 0.849±0.039             \\
            DiCNN~\cite{hePansharpeningDetailInjection2019}           & 0.0947±0.0145          & 0.1067±0.0210          & 0.809±0.031             \\
            FusionNet~\cite{dengDetailInjectionBasedDeep2021}         & 0.0572±0.0182          & 0.0522±0.0088          & 0.894±0.021             \\
            DCFNet~\cite{wuDynamicCrossFeature2021}                   & 0.0469±0.0150          & 0.1239±0.0269          & 0.835±0.016             \\
            MMNet~\cite{mmnet}                                        & 0.0768±0.0257          & \textbf{0.0374±0.0201} & 0.889±0.041             \\
            LAGConv~\cite{jinLAGConvLocalContextAdaptive2022}         & 0.0859±0.0237          & 0.0676±0.0136          & 0.852±0.018             \\
            HMPNet~\cite{hmpnet}                                      & 0.1838±0.0542          & 0.0793±0.0245          & 0.753±0.065             \\
            \textbf{Proposed}                                         & \uline{0.0370±0.0129}  & \uline{0.0499±0.0092}  & \textbf{0.915±0.012}
        \end{tblr}
    }
\end{table}

\begin{table}
    \centering
    \caption{Result benchmark on the GF2 dataset with 20 full-resolution samples. \textbf{Bold}: best, \uline{underline}: second best.}
    \label{tab:gf2full}
    \resizebox{\linewidth}{!}{%
        \begin{tblr}{
            column{even} = {c},
            column{3} = {c},
            hline{1-2,8,16-17} = {-}{},
                }
            \textbf{Method}                                           & $D_\lambda\downarrow$  & $D_s\downarrow$        & \textbf{HQNR}$\uparrow$ \\
            EXP~\cite{EXP}                                            & \uline{0.0180±0.0081}  & 0.0957±0.0209          & 0.888±0.023             \\
            TV~\cite{palssonNewPansharpeningAlgorithm2014}            & 0.0346±0.0137          & 0.1429±0.0282          & 0.828±0.035             \\
            MTF-GLP-FS~\cite{vivoneFullScaleRegressionBased2018}      & 0.0553±0.0430          & 0.1118±0.0226          & 0.839±0.044             \\
            BDSD-PC~\cite{vivoneRobustBandDependentSpatialDetail2019} & 0.0759±0.0301          & 0.1548±0.0280          & 0.781±0.041             \\
            CVPR19~\cite{fuVariationalPanSharpeningLocal2019}         & 0.0307±0.0127          & \textbf{0.0622±0.0101} & 0.909±0.017             \\
            LRTCFPan~\cite{LRTCFPan}                                  & 0.0325±0.0269          & 0.0896±0.0141          & 0.881±0.023             \\
            PNN~\cite{masiPansharpeningConvolutionalNeural2016a}      & 0.0317±0.0286          & 0.0943±0.0224          & 0.877±0.036             \\
            PanNet~\cite{yangPanNetDeepNetwork2017}                   & \textbf{0.0179±0.0110} & 0.0799±0.0178          & 0.904±0.020             \\
            DiCNN~\cite{hePansharpeningDetailInjection2019}           & 0.0369±0.0132          & 0.0992±0.0131          & 0.868±0.016             \\
            FusionNet~\cite{dengDetailInjectionBasedDeep2021}         & 0.0350±0.0124          & 0.1013±0.0134          & 0.867±0.018             \\
            DCFNet~\cite{wuDynamicCrossFeature2021}                   & 0.0240±0.0115          & 0.0659±0.0096          & \uline{0.912±0.012}     \\
            MMNet~\cite{mmnet}                                        & 0.0443±0.0298          & 0.1033±0.0129          & 0.857±0.027             \\
            LAGConv~\cite{jinLAGConvLocalContextAdaptive2022}         & 0.0284±0.0130          & 0.0792±0.0136          & 0.895±0.020             \\
            HMPNet~\cite{hmpnet}                                      & 0.0819±0.0499          & 0.1146±0.0126          & 0.813±0.049             \\
            \textbf{Proposed}                                         & 0.0194±0.0101          & \uline{0.0630±0.0094}  & \textbf{0.919±0.011}
        \end{tblr}
    }
\end{table}

\subsection{Additional Results}

\label{sec:AdditionalResults}

% \vspace{-0.2cm}

\begin{figure*}
  \centering
  \includegraphics[width=1\linewidth]{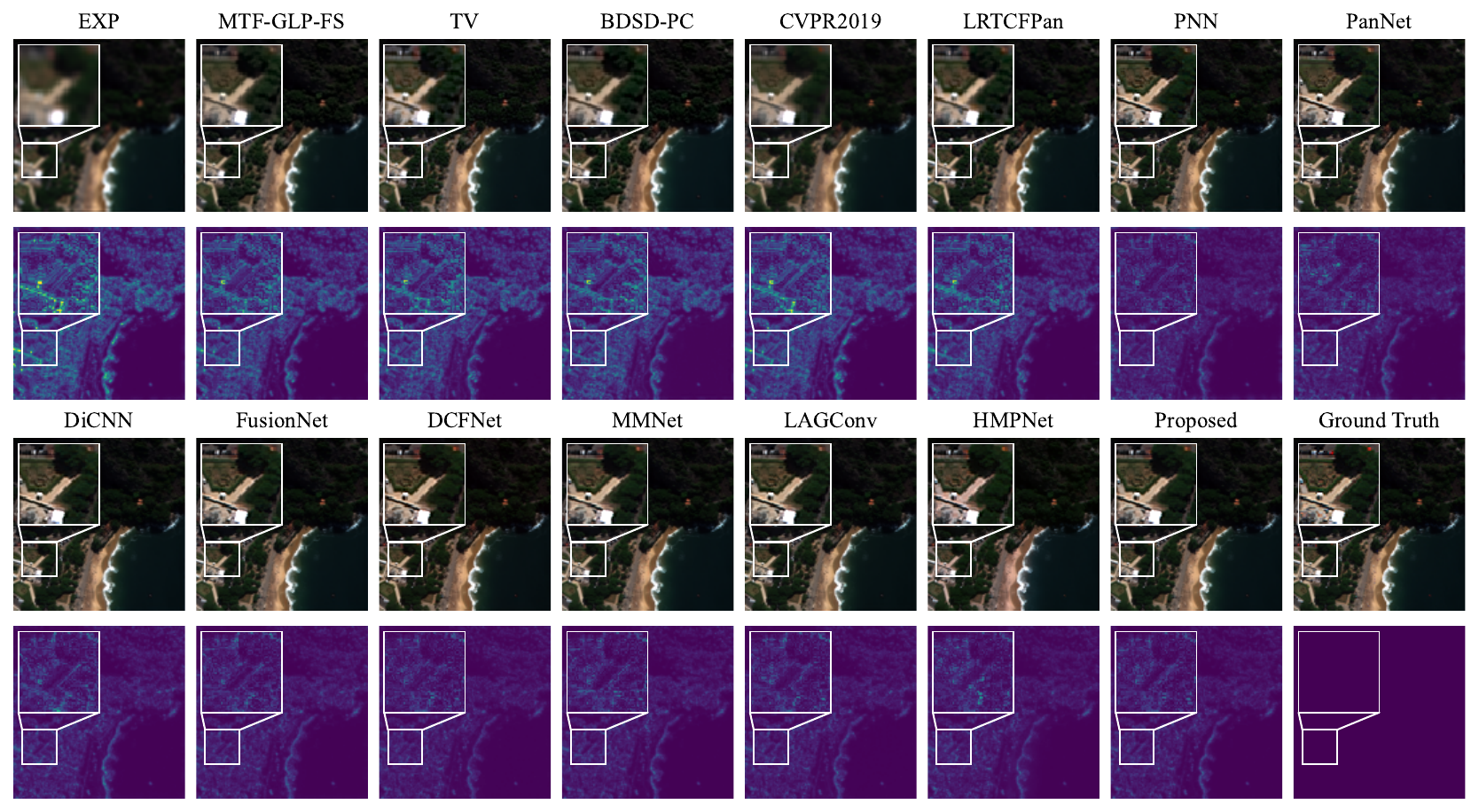}
  \caption{Qualitative result comparison between benchmarked methods on the sample image from WV3 reduced-resolution dataset. The first row presents RGB outputs, while the second row shows the residual compared to the ground truth. }
  \label{fig:WV3_Reduced5}
\end{figure*}

\begin{figure*}
  \centering
  \includegraphics[width=1\linewidth]{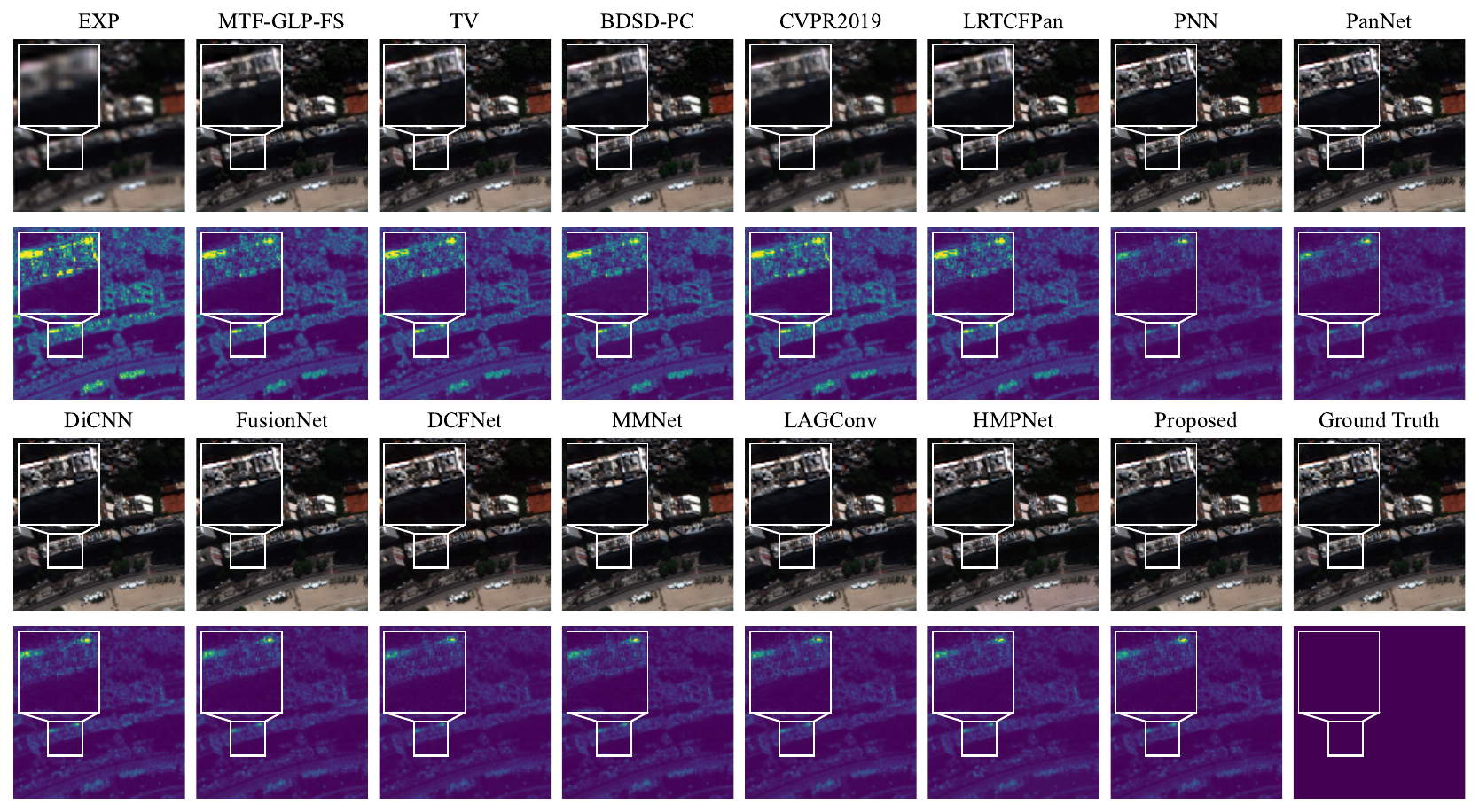}
  \caption{Qualitative result comparison between benchmarked methods on the sample image from WV3 reduced-resolution dataset.}
  \label{fig:WV3_Reduced10}
\end{figure*}

\begin{figure*}
  \centering
  \includegraphics[width=1\linewidth]{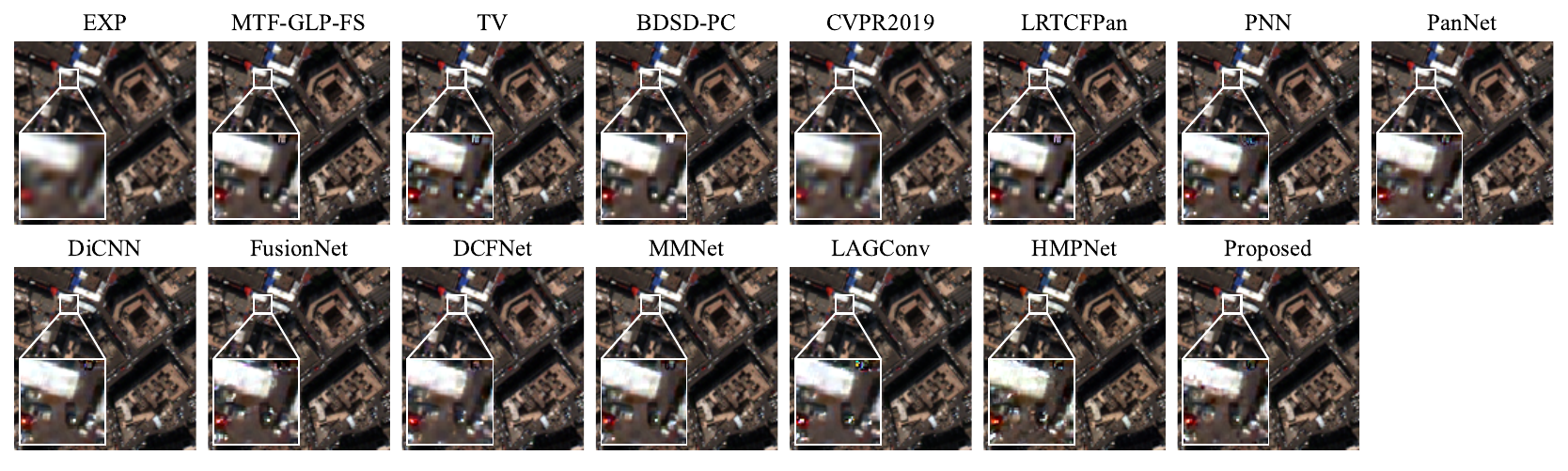}
  \caption{Qualitative result comparison between benchmarked methods on the sample image from the WV3 full-resolution dataset. }
  \label{fig:WV3_Full11}
  \vspace{-0.4cm}
\end{figure*}

\begin{figure*}
  \centering
  \includegraphics[width=1\linewidth]{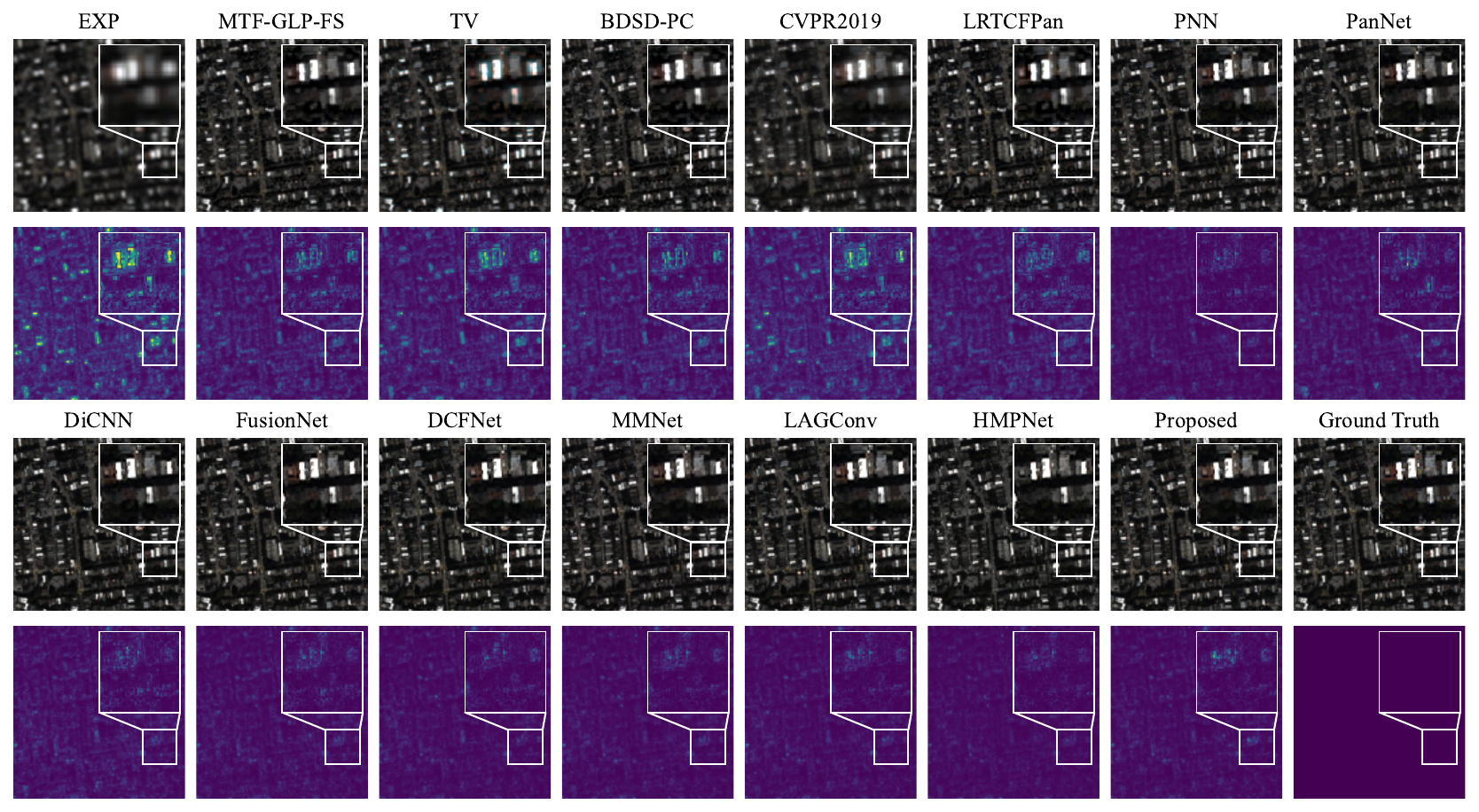}
  \caption{Qualitative result comparison between benchmarked methods on the sample image from the QB reduced-resolution dataset.}
  \label{fig:QB_Reduced8}
  \vspace{-0.4cm}
\end{figure*}

\begin{figure*}
  \centering
  \includegraphics[width=1\linewidth]{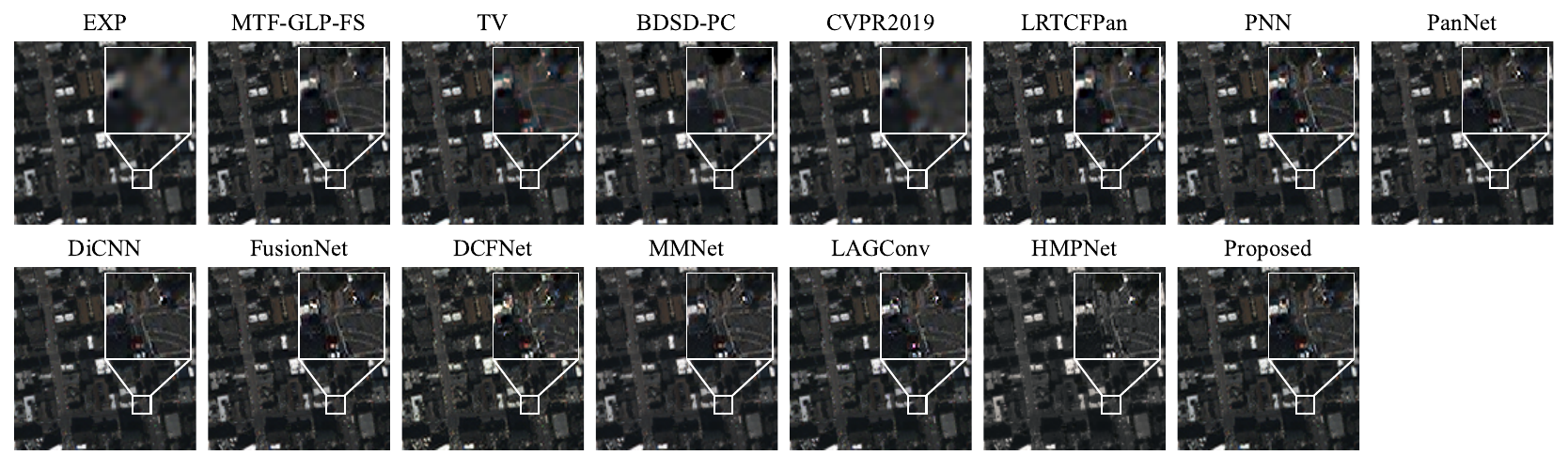}
  \caption{Qualitative result comparison between benchmarked methods on the sample image from the QB full-resolution dataset. }
  \label{fig:QB_Full10}
\end{figure*}

\begin{figure*}
  \centering
  \includegraphics[width=1\linewidth]{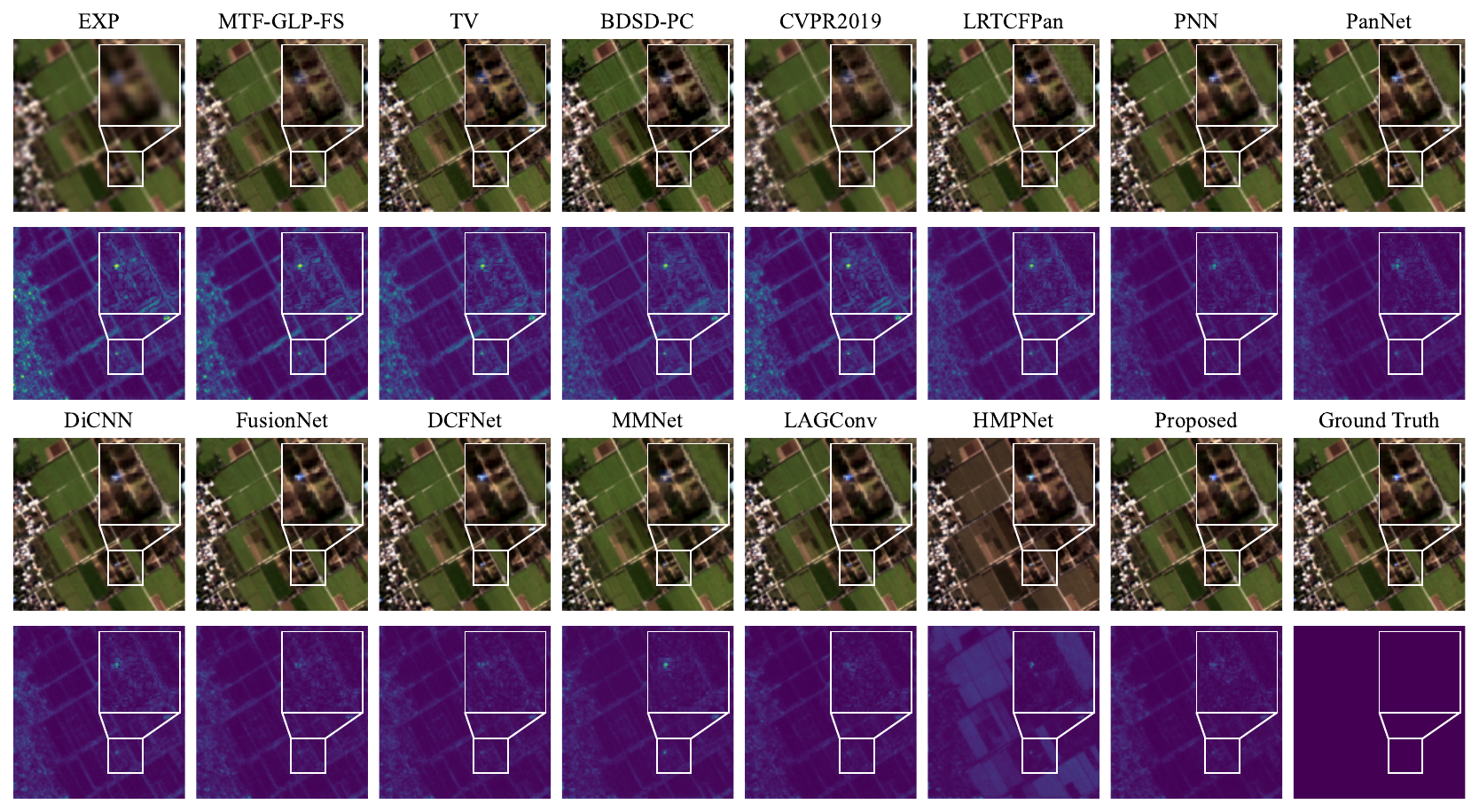}
  \caption{Qualitative result comparison between benchmarked methods on the sample image from the GF2 reduced-resolution dataset.}
  \label{fig:GF2_Reduced19}
  \vspace{-0.4cm}
\end{figure*}

\begin{figure*}
  \centering
  \includegraphics[width=1\linewidth]{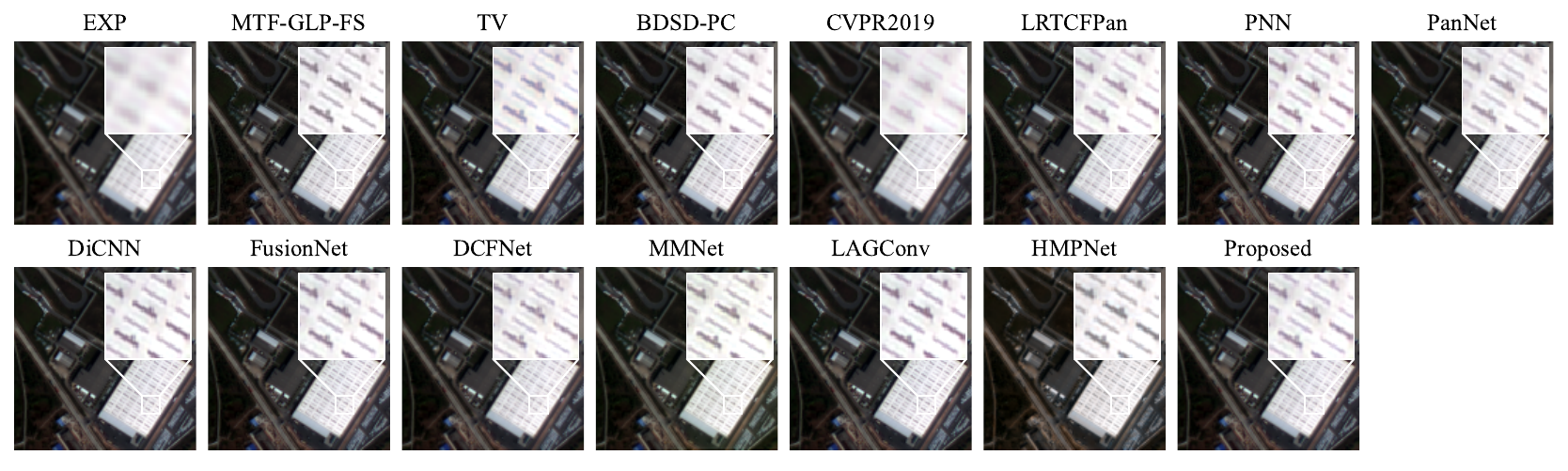}
  \caption{Qualitative result comparison between benchmarked methods on the sample image from the GF2 full-resolution dataset. }
  \label{fig:GF2_Full0}
  \vspace{-0.4cm}
\end{figure*}

\begin{figure*}
  \centering
  \includegraphics[width=1\linewidth]{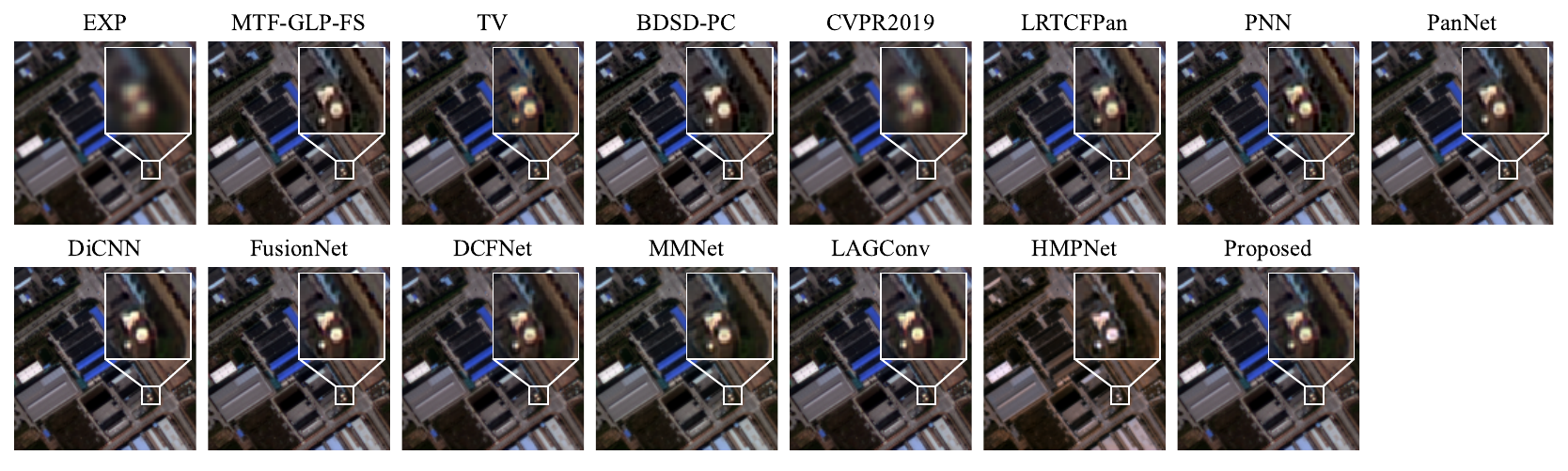}
  \caption{Qualitative result comparison between benchmarked methods on the sample image from the GF2 full-resolution dataset. }
  \label{fig:GF2_Full10}
\end{figure*}

\cref{tab:qbfull,tab:gf2full} showcase performance benchmarks on the full-resolution QB and GF2 datasets. The HQNR metric~\cite{hqnr} is an improvement upon the QNR metric. Combining assessments of both spatial and spectral consistency, HQNR provides a comprehensive reflection of the image-fusion effectiveness of different methods. It is considered one of the most important metrics on full-resolution datasets. In \cref{fig:WV3_Reduced5,fig:WV3_Reduced10,fig:WV3_Full11,fig:QB_Reduced8,fig:QB_Full10,fig:GF2_Reduced19,fig:GF2_Full0,fig:GF2_Full10}, we present visual output comparisons across various methods on sample images from the WV3, QB and GF2 datasets, including residuals between outputs and ground truth for reduced-resolution samples. The comparative analysis highlights that, overall, CANNet produces results closely aligned with the ground truth. Leveraging self-similarity information, CANNet excels in handling repetitive texture areas, surpassing the performance of previous methods, as shown in \cref{fig:GF2_Full0}. Also, CANNet exhibits adaptive processing in detail-rich edge regions, resulting in more realistic and accurate outcomes.

\FloatBarrier
% \clearpage
% {
%     \small
%     \bibliographystyle{ieee}
%     \bibliography{main}
% }

\end{document}